\documentclass[12pt]{article}
\usepackage[margin=1in]{geometry}
\usepackage{amsmath,amssymb,amsthm,mathtools}
\usepackage{enumitem}
\usepackage{graphicx,booktabs,float}
\usepackage[section]{placeins}
\graphicspath{{figures/}{./}}
\usepackage[colorlinks=true,citecolor=blue,linkcolor=blue,urlcolor=blue]{hyperref}
\usepackage[authoryear,round]{natbib}
\usepackage{url}
\newtheorem{proposition}{Proposition}
\newtheorem{theorem}{Theorem}

\newtheorem{lemma}{Lemma}
\theoremstyle{remark}

\title{Task-Calibrated Random Features \\via Influence-Based Error Cancellation}
\author{Shinto Eguchi}
\date{\null}

\begin{document}
\maketitle

\begin{abstract}
Random features approximate kernel integrals by Monte Carlo, yet only selected frequency-sampling errors affect a downstream task. We propose Task-Calibrated Random Features (TCRF), which balance frequency moments constructed from task-specific influence using weights shared across kernel entries. Closed-form weights coincide with classical minimum-distance calibration; the contribution is a reusable, task-derived calibration space. We show that first-order variance equals the residual variance after projection onto this space. For kernel ridge regression, complete cancellation yields a non-Gaussian second-order limit. For kernel Stein discrepancy, a subset identity and disjoint-block variance decomposition eliminate the linear component of pilot noise. Positive exponential-tilting weights preserve the positive-semidefinite feature representation and support higher feasible calibration ranks. Paired experiments show improved approximation to exact kernel ridge regression at larger frequency budgets and to the exact empirical Stein discrepancy under a fixed alternative, even after accounting for a frequency-equivalent pilot budget. Gains are smaller with cross-validated ridge, landmark pilots require conservative rank selection, and near the Stein null allocating the pilot budget to additional ordinary frequencies is preferable.
\end{abstract}

\section{Introduction}

Random Fourier features replace a shift-invariant kernel integral by a finite Monte Carlo average over sampled frequencies \citep{rahimi2007}. Their computational appeal is clear, but the approximation error is not equally important in every direction. The error that propagates to a downstream task often occupies only a few of the directions in which a finite frequency sample deviates from its spectral law. This paper asks whether those task-relevant directions can be identified in advance and cancelled directly on the sampled frequencies, without changing the sampling distribution. The baseline construction uses a closed-form weight and no inner optimisation; a positive exponential-tilting extension is investigated separately.

Our point of departure is the estimated-sampler phenomenon of \citet{henmi2007}. There, even when the true sampling distribution of an importance sampler is known, re-estimating its nuisance parameter on the same Monte Carlo sample reduces the variance, because the score equation removes the component of the target fluctuation that lies in the score space. If the target fluctuation lies entirely in the span of the score, the first-order asymptotic variance is zero. We use this zero- and near-zero-variance geometry as a design principle for random features, but we do not estimate a sampler parameter. What we retain is the projection: the space onto which the Monte Carlo fluctuation is projected can be designed, and when it is designed from the downstream task it removes precisely the error that the task sees.

We propose Task-Calibrated Random Features (TCRF). Here calibration means exact moment balancing of the sampled frequencies, not probabilistic calibration of predictive probabilities. Given a few frequency functions with known mean zero, the closed-form version assigns each sampled frequency one weight, shared by all kernel entries and test points, so that the empirical moments vanish exactly. The weights require only a sample mean and covariance. The first-order error is the task fluctuation projected onto the orthogonal complement of the calibration space, so complete containment yields zero first-order variance and approximate containment yields near-zero variance. We also consider KL-based exponential tilting (ET), which maintains positive weights on feasible calibration problems and can enlarge the practically usable calibration space.

In kernel ridge regression the relevant frequency-wise quantity is the contribution of a single frequency to the prediction error, not the error of an isolated kernel entry. We construct a low-rank calibration space from the principal directions of this prediction influence, show that it minimises the residual variance at a given rank, and derive the second-order regime that remains when the first-order influence is cancelled completely: the error then decreases in inverse proportion to the number of frequencies, rather than its square root, and its limit is a signed quadratic form in Gaussian kernel errors. In the kernel Stein discrepancy the frequency integral is itself the computational target. A subset identity shows that the centred integrand of a random subset of the data recovers, in expectation, the full-data zero-variance direction up to a single isotropic scale term. Disjoint-block pilots remove the linear component of subset noise exactly; under a fixed alternative, the subset size required for fixed relative precision need not grow with the sample size, so the pilot cost scales linearly with that size.

The closed-form weights themselves are not new. They are exactly the minimum-distance calibration weights of the regression or generalised regression form of \citet{deville1992}, and for a scalar Monte Carlo target the TCRF estimator is algebraically identical to a same-sample regression control-variate estimator \citep{lavenberg1982,nelson1990,davidson1992,portier2019}. Our contribution is to design the controls from downstream learning influence and to reuse the resulting common weights as a random-feature representation. When the selected weights are nonnegative, they can be absorbed into the feature coordinates, so the weighted kernel remains positive semidefinite with unit diagonal and the squared Stein discrepancy remains nonnegative. The ET extension enforces positive weights on feasible instances, at an additional numerical optimisation cost, and is therefore a complementary implementation rather than a change to the projection theorems. It is related to exponential-tilting calibration in survey sampling and entropy balancing in observational studies \citep{kim2010,hainmueller2012}, but balances sampled frequencies rather than survey or treatment groups. A generic signed correction applied after the fact need not preserve these structures, a representational issue already highlighted in the moment-matching literature \citep{shen2017}.

A substantial literature reduces random-feature error by changing the frequency design. Moment matching removes low-order empirical frequency fluctuations \citep{shen2017}; orthogonal random features replace independent Gaussian directions by suitably scaled orthogonal ones \citep{yu2016}; and quasi-Monte Carlo constructions use low-discrepancy sequences \citep{yang2014,huang2024}. These methods improve the spectral design without using the downstream influence that is central here. Kernel quadrature and random feature expansions are linked by \citet{bach2017}, and \citet{rudi2017} study learning guarantees for random features; our focus is conditional Monte Carlo error in a fixed downstream computation. A second line of work adapts the sampling distribution to the learning problem: \citet{avron2017} analyse random features for kernel ridge regression through spectral approximation and leverage-based sampling, and \citet{li2019} develop a unified risk analysis with a data-dependent scheme based on ridge leverage scores. These methods change which frequencies are sampled, whereas TCRF reweights a given frequency sample so as to annihilate selected task-relevant moments; the two approaches are complementary. On the downstream side, \citet{yao2023} estimate how random-feature error propagates into particular learning tasks and use this information for adaptive computation; we instead cancel the first-order influence directly. For the Stein discrepancy, random-feature and importance-sampling approximations were studied by \citet{huggins2018}, while \citet{jitkrittum2017} design features for a linear-time goodness-of-fit test. Our construction targets the frequency-integration error of the existing empirical discrepancy, rather than a new discrepancy or test.

The zero-variance principle has a broader history in variance-reduced
Monte Carlo computation, including zero-variance estimators for
quantum Monte Carlo and Markov chain Monte Carlo
\citep{assaraf1999zero,mira2013zero} and control-functionals
integration \citep{oates2017control}. Those works modify an estimator
for a specified integration target; our emphasis is on extracting a
downstream learning influence to construct common frequency weights
that retain the random-feature representation. Closely related
frequency-adaptive approaches include data-driven Stein-effect
shrinkage of Fourier-feature weights \citep{chang2017stein},
data-dependent compression or selection of random features
\citep{agrawal2019compression}, and learning kernels through their
random-feature distributions \citep{sinha2016kernels}.
Unlike these approaches, the central construction here imposes exact
balance on prespecified, analytically centred task directions without
learning a new spectral law; ET changes only the calibration
divergence and finite-sample positive-weight implementation.

Section~\ref{sec:framework} recalls the estimated-sampler geometry and
introduces the two downstream computations. Section~\ref{sec:main}
develops the closed-form calibration principle and presents two
central results: the second-order limit for kernel ridge regression
and the disjoint-block variance decomposition for the kernel Stein
discrepancy. Section~\ref{sec:numerical} reports paired Monte Carlo
experiments, and Section~\ref{sec:discussion} discusses the scope
and limitations of the method. Additional proofs and derivations
appear in the Appendix, while additional numerical results are given in Appendix~\ref{supp:numerical}.

\section{Framework}
\label{sec:framework}

\subsection{Cancellation by the estimated sampler}
\label{sec:estimated-sampler}

Consider the integral
\[
\theta=\int f(x)\,dx
\]
with sampling density $p(x;\eta)$, and let $X_1,\ldots,X_m$ be generated from $p(x;\eta_0)$ for a known true value $\eta_0$. The ordinary importance sampling estimator is
\[
\tilde\theta=\frac{1}{m}\sum_{j=1}^m \frac{f(X_j)}{p(X_j;\eta_0)}.
\]
Write
\[
u(X)=\frac{f(X)}{p(X;\eta_0)}-\theta,
\quad
s(X)=\left.\frac{\partial}{\partial\eta}\log p(X;\eta)\right|_{\eta=\eta_0}.
\]
Instead of using the true $\eta_0$, let $\hat\eta$ be the maximum likelihood estimator computed from the same sample and set
\[
\hat\theta=\frac{1}{m}\sum_{j=1}^m \frac{f(X_j)}{p(X_j;\hat\eta)}.
\]
A first-order expansion (derived in Appendix~\ref{supp:estimated-sampler}) gives
\[
\sqrt{m}(\hat\theta-\theta)
=
\frac{1}{\sqrt m}\sum_{j=1}^m
\left(
 u(X_j)-\mathbb{E}(u s^{\top})G^{-1}s(X_j)
\right)
+o_p(1),
\]
where $G=\mathbb{E}[s(X)s(X)^\top]$.
Thus the second term is the projection of $u$ onto the span of the score in $L_2(p_{\eta_0})$. Consequently, the first-order asymptotic variances satisfy
\[
\operatorname{avar}(\tilde\theta)
-
\operatorname{avar}(\hat\theta)
=
\mathbb{E}(u s^{\top})G^{-1}\mathbb{E}(su)
\geq 0.
\]
The benefit of the estimated sampler does not come from the accuracy with which the nuisance parameter is estimated, but from the fact that the score equation absorbs part of the Monte Carlo fluctuation of interest. Writing $\mathcal S$ for the span of the score and $\Pi_{\mathcal S}$ for the orthogonal projection onto it in $L_2(p_{\eta_0})$, the asymptotic variance of the estimated-sampler estimator is
\[
\operatorname{avar}(\hat\theta)
=
\left\|(I-\Pi_{\mathcal S})u\right\|_{L_2(p_{\eta_0})}^2.
\]
Therefore, the first-order asymptotic variance is exactly zero if and only if $u$ lies in the linear span $\mathcal S$. What matters is not the direct construction of a perfect zero-variance sampler, but the inclusion of the target fluctuation in the cancellation space. Complete inclusion gives zero variance, and approximate inclusion leaves only the magnitude of the orthogonal residual as variance. This zero- and near-zero-variance viewpoint is the point of contact with the learning problems considered below.

The use of the same sample is essential. If $\hat\eta$ were computed from an independent sample, the cancellation arising from the covariance between $u$ and the score fluctuation would not occur. The score equation
\[
\sum_{j=1}^m s(X_j;\hat\eta)=0
\]
on the same sample can be read as implicitly determining an empirical regression or control-variate coefficient. In this sense the structure found by \citet{henmi2007} is not merely a correction to importance sampling, but a self-calibration mechanism that projects the target fluctuation onto the nuisance tangent space on a single Monte Carlo sample. The method proposed below uses only this projection geometry, and computes neither the sampler parameter nor its estimate.

\subsection{Frequency approximation in KRR and KSD}

\paragraph{Gaussian features and moment matching.}

For the Gaussian kernel with bandwidth $\sigma>0$, let
\[
k_\sigma(x,x')=\exp\left(-\frac{\|x-x'\|^2}{2\sigma^2}\right),
\quad \delta=\frac{x-x'}{\sigma}.
\]
For brevity, write $k=k_\sigma$ when the bandwidth is fixed.
The spectral representation with $Z\sim N(0,I_d)$ entails
$
k(x,x')=\mathbb{E}_Z\!\left[\cos(Z^\top\delta)\right].
$
One frequency defines the real two-dimensional feature block
\[
\phi_Z(x)=
\begin{pmatrix}\cos(Z^\top x/\sigma)\\ \sin(Z^\top x/\sigma)\end{pmatrix},
\quad
\phi_Z(x)^\top\phi_Z(x')=\cos(Z^\top\delta).
\]
For independent $Z_1,\ldots,Z_m\sim N(0,I_d)$, the ordinary random Fourier
feature (RFF) kernel approximation is
\[
\hat k_{\mathrm{RFF}}(\delta)
=\frac1m\sum_{j=1}^m\cos(Z_j^\top\delta).
\]
To distinguish two existing methods of reducing its Monte Carlo error, put
\[
\bar Z=\frac1m\sum_{j=1}^m Z_j,\quad
S_m=\frac1m\sum_{j=1}^m(Z_j-\bar Z)(Z_j-\bar Z)^\top,
\]
and suppose that $S_m$ is nonsingular. Gaussian moment matching (MM)
centres and whitens these same sampled frequencies, giving
\[
\widetilde Z_j=S_m^{-1/2}(Z_j-\bar Z),\quad
\hat k_{\mathrm{MM}}(\delta)
=\frac1m\sum_{j=1}^m\cos(\widetilde Z_j^\top\delta).
\]
By contrast, the estimated-sampler (ES) construction estimates the mean
and covariance of a Gaussian sampling density from these frequencies and
reweights each original frequency. With $q_{\mu,\Sigma}$ the density of
$N(\mu,\Sigma)$ and $q_0=q_{0,I_d}$, define
\[
\hat k_{\mathrm{ES}}(\delta)
=\frac1m\sum_{j=1}^m
\frac{q_0(Z_j)}{q_{\bar Z,S_m}(Z_j)}
\cos(Z_j^\top\delta).
\]
ES and MM operate differently---the former reweights frequencies and the
latter transforms them---but, for fixed $d$ and $\delta$, their first-order
asymptotic expansions coincide. In particular,
$\hat k_{\mathrm{ES}}(\delta)-\hat k_{\mathrm{MM}}(\delta)
=o_p(m^{-1/2})$. Both remove the same Gaussian covariance-score
component of the frequency fluctuation to first order; by symmetry the
Gaussian mean-score component vanishes at this order. Hence Gaussian
sampler re-estimation alone is not our contribution. What matters for the
proposed method is to select cancellation directions from the downstream
task rather than to fix them a priori as Gaussian scores.
Detailed expansions and covariance computations appear in
Appendix~\ref{supp:mm}; see also \citet{shen2017} for Gaussian
moment matching.

\paragraph{Kernel ridge regression.}
The first task we consider is kernel ridge regression (KRR).
For large training samples, exact KRR requires $O(n^2)$ memory to store the
$n\times n$ Gram matrix and $O(n^3)$ arithmetic operations for a direct
factorisation. Random Fourier features replace the kernel with a
low-dimensional feature representation, so that a much smaller linear
system can be solved when the number of features is far below $n$.
Our immediate purpose is to quantify the Monte Carlo error of this
computational approximation, before introducing any task calibration.

Fix training data $(X_i,y_i)_{i=1}^n$, with $X_i\in\mathbb R^d$ and
$y=(y_1,\ldots,y_n)^\top$, and choose $\lambda>0$. Define the exact
kernel Gram matrix and test-to-training kernel vector by
\[
K=\bigl(k(X_i,X_{i'})\bigr)_{i,i'=1}^n\in\mathbb R^{n\times n},
\quad
k_x=\bigl(k(x,X_i)\bigr)_{i=1}^n\in\mathbb R^n.
\]
Write $R=(K+n\lambda I_n)^{-1}$ and $\alpha=Ry$. The exact KRR predictor is
\[
f_K(x)=k_x^\top\alpha.
\]
For the $m$ sampled frequencies defined above, stack the paired cosine
and sine features as
\[
\Phi_m(x)=\frac1{\sqrt m}
\begin{pmatrix}
\phi_{Z_1}(x)\\ \vdots\\ \phi_{Z_m}(x)
\end{pmatrix}\in\mathbb R^{2m}.
\]
The training feature matrix is
\[
F=\begin{pmatrix}
\Phi_m(X_1)^\top\\ \vdots\\ \Phi_m(X_n)^\top
\end{pmatrix}\in\mathbb R^{n\times 2m}.
\]
The random-feature approximations to the training Gram matrix and the
test-to-training kernel vector are
\[
\widehat K=FF^\top,\quad \hat k_x=F\Phi_m(x).
\]
They satisfy $\mathbb{E}_Z[\widehat K]=K$ and
$\mathbb{E}_Z[\hat k_x]=k_x$, where the expectation is over the
sampled frequencies conditional on the training data.

The random-feature KRR predictor is
\[
\hat f(x)
=\hat k_x^\top(\widehat K+n\lambda I_n)^{-1}y.
\]
It can also be computed in feature space as
\[
\hat f(x)
=\Phi_m(x)^\top(F^\top F+n\lambda I_{2m})^{-1}F^\top y.
\]
The second representation follows from $\widehat K=FF^\top$,
$\hat k_x=F\Phi_m(x)$, and the matrix identity
\[
F^\top(FF^\top+n\lambda I_n)^{-1}
=(F^\top F+n\lambda I_{2m})^{-1}F^\top,
\]
which holds for $\lambda>0$.
When $2m\ll n$, a dense feature-space fit costs
$O(nmd+nm^2+m^3)$ operations, including feature construction,
and $O(nm+m^2)$ memory if the feature matrix is stored.
The feature Gram matrix can also be accumulated in batches.
These are costs for the random-feature fit, not for constructing
task-influence calibration directions.

The solver used in a numerical experiment need not use the smaller
feature-space system. The main KRR and landmark experiments in
Section~\ref{sec:numerical} form the weighted $n\times n$ Gram matrix
and solve the $n$-dimensional dual system for all frequency budgets,
even when $2m<n$. They are controlled accuracy comparisons, not timing
benchmarks for a feature-space implementation. By contrast, the
large-sample benchmark in Appendix~\ref{sec:supp-large-krr} chooses the smaller
of the $n$-dimensional dual and $2m$-dimensional primal systems.
The full-influence approach also requires the exact kernel matrices
and exact KRR solution for its pilot directions; the landmark construction instead uses a reduced exact problem,
and the separate Nystr\"om--Taylor approximation has its own centring cost.
The feature-space complexity alone therefore does not establish an
end-to-end speed-up of the full-influence procedure.

The gain in computation comes with a frequency-sampling approximation
error relative to the exact predictor $f_K(x)$. To express its first-order
perturbation, define the contribution of one frequency $Z\sim N(0,I_d)$
to the training Gram matrix by
\[
(A_Z)_{ii'}=\cos\!\left(\frac{Z^\top(X_i-X_{i'})}{\sigma}\right),
\quad i,i'=1,\ldots,n.
\]
Likewise, define its contribution to the test-to-training kernel vector by
\[
(a_Z(x))_i=\cos\!\left(\frac{Z^\top(x-X_i)}{\sigma}\right),
\quad i=1,\ldots,n.
\]
Consequently,
$\widehat K=m^{-1}\sum_{j=1}^m A_{Z_j}$ and
$\hat k_x=m^{-1}\sum_{j=1}^m a_{Z_j}(x)$, with
$\mathbb{E}_Z[A_Z]=K$ and $\mathbb{E}_Z[a_Z(x)]=k_x$.
For fixed training data, $n$ and $\lambda$, the first-order expansion is
\[
\hat f(x)-f_K(x)
=\frac1m\sum_{j=1}^m b_x(Z_j)+o_p(m^{-1/2}).
\]
Here the prediction influence of a single frequency is
\[
b_x(Z)=(a_Z(x)-k_x)^\top\alpha
-k_x^\top R(A_Z-K)\alpha.
\]
It is centred under the spectral law:
\[
\mathbb{E}_Z[b_x(Z)]=0.
\]
This quantity $b_x(Z)$ measures how a single sampled frequency
perturbs the KRR prediction, not merely one entry of the Gram matrix.
It is therefore the appropriate task-dependent fluctuation to examine.
The derivation and its rank-two implementation are given in
Appendix~\ref{supp:krr}.

\paragraph{Kernel Stein discrepancy.}
The second task is the kernel Stein discrepancy (KSD). Let $p$ be a
target density on $\mathbb R^d$ with score $s_p(x)=\nabla_x\log p(x)$,
let $D=\{X_1,\ldots,X_n\}$ be the data, and consider the Gaussian kernel
\[
k_\sigma(x,y)=\exp\left(-\frac{\|x-y\|^2}{2\sigma^2}\right)
=\mathbb{E}_\omega[\cos(\omega^\top(x-y))],
\quad \omega\sim N(0,\sigma^{-2}I_d).
\]
Thus $\omega=Z/\sigma$ for the standard Gaussian frequency
$Z\sim N(0,I_d)$ used in the KRR part.
For the Langevin Stein operator, define the real-valued Stein feature
\[
\psi_\omega(x)
=\begin{pmatrix}
s_p(x)\cos(\omega^\top x)-\omega\sin(\omega^\top x)\\
s_p(x)\sin(\omega^\top x)+\omega\cos(\omega^\top x)
\end{pmatrix}
\in\mathbb R^{2d}.
\]
The Stein kernel $h_p$ associated with $k_\sigma$ admits the real Fourier representation
\[
h_p(x,y)=\mathbb{E}_\omega
[\psi_\omega(x)^\top\psi_\omega(y)].
\]
Its empirical V-statistic defines the squared kernel Stein discrepancy
used for goodness of fit \citep{liu2016,chwialkowski2016}:
\[
\theta_D
=\frac1{n^2}\sum_{i=1}^n\sum_{j=1}^n h_p(X_i,X_j)
=\mathbb{E}_\omega[G_D(\omega)],
\]
where the frequency integrand is
\[
G_D(\omega)
=\left\|\frac1n\sum_{i=1}^n\psi_\omega(X_i)\right\|^2.
\]
The random-feature discrepancy of \citet{huggins2018} is the average
of $G_D$ over sampled frequencies. For the fixed dataset $D$ put
\[
u_D(\omega)=G_D(\omega)-\theta_D.
\]
In contrast with kernel ridge regression, the frequency integral is itself the computational target, so no linearisation is involved. The exact statistic requires $n^2$ evaluations of the Stein kernel, at cost of order $n^2d$, whereas the random-feature computation with $m$ frequencies costs order $mnd$. The gain is substantial when $m\ll n$, provided that the frequency error is small, and the integrand $u_D$ plays here exactly the role of the target fluctuation $u$ of the estimated-sampler problem.

\section{Main result}
\label{sec:main}

\subsection{Closed-form task calibration}

Let $q_0$ be the base spectral distribution (the law of $Z$ for kernel ridge regression and of $\omega$ for the Stein discrepancy), and let
$Z_1,\ldots,Z_m$ be independent and identically distributed according
to $q_0$. For a scalar integrand $h$, define the Monte Carlo target
$
\theta=\mathbb{E}_{q_0}[h(Z)].
$
Define its centred fluctuation by
$
u(Z)=h(Z)-\theta.
$
Next, choose $r$ calibration functions and collect them into the vector
$
c(Z)=(c_1(Z),\ldots,c_r(Z))^{\top}.
$
Their expectations are assumed known and equal to zero:
$
\mathbb{E}_{q_0}[c(Z)]=0.
$
Define the sample mean of the calibration functions and the empirical covariance matrix by
\[
\bar c_m=\frac1m\sum_{j=1}^m c(Z_j),\quad
\widehat\Sigma_c
=
\frac1m\sum_{j=1}^m
(c(Z_j)-\bar c_m)(c(Z_j)-\bar c_m)^{\top}.
\]
We define \emph{Task-Calibrated Random Features} (TCRF) as the
random-feature procedure that assigns common, task-calibrated weights to
the sampled frequencies. When $\widehat\Sigma_c$ is nonsingular, its
closed-form frequency weights are
\begin{equation}
w_j
=
1-(c(Z_j)-\bar c_m)^{\top}
\widehat\Sigma_c^{-1}\bar c_m .
\label{eq:closed-form-weight}
\end{equation}
The associated TCRF estimator of the scalar target is
\[
\hat\theta_{\mathrm{TCRF}}
=\frac1m\sum_{j=1}^m w_jh(Z_j).
\]
The TCRF weights require no sampler parameter, dual parameter or iterative
optimisation. Direct substitution gives the exact finite-sample balance
relations
\begin{equation}
\frac1m\sum_{j=1}^m w_j=1,
\quad
\frac1m\sum_{j=1}^m w_jc(Z_j)=0.
\label{eq:exact-balance}
\end{equation}
The weights in \eqref{eq:closed-form-weight} are a special case of classical calibration weights. With equal base weights and known auxiliary mean zero, they solve the quadratic minimum-distance problem
\begin{equation}
\min_{w_1,\ldots,w_m}
\frac1m\sum_{j=1}^m(w_j-1)^2
\quad\text{subject to}\quad
\frac1m\sum_{j=1}^m w_j=1,\quad
\frac1m\sum_{j=1}^m w_jc(Z_j)=0,
\label{eq:min-distance}
\end{equation}
which is the chi-square calibration construction of \citet{deville1992} specialised to equal base weights with the intercept included among the auxiliary variables. Indeed, the first-order condition of \eqref{eq:min-distance} makes $w_j-1$ an affine function of $c(Z_j)$; the constraint $m^{-1}\sum_jw_j=1$ centres this function at $\bar c_m$, and the balance constraint then fixes its coefficient vector at $-\widehat\Sigma_c^{-1}\bar c_m$, which is \eqref{eq:closed-form-weight}. Strict convexity of the objective makes this solution unique. Because the weights are affine in $c(Z_j)-\bar c_m$, which has empirical mean zero, substituting them into $\hat\theta_{\mathrm{TCRF}}$ gives, with
\[
\bar h_m=\frac1m\sum_{j=1}^m h(Z_j),
\quad
\widehat{\operatorname{Cov}}(h,c)
=\frac1m\sum_{j=1}^m
(h(Z_j)-\bar h_m)(c(Z_j)-\bar c_m),
\]
the finite-sample identity
\begin{equation}
\hat\theta_{\mathrm{TCRF}}
=
\bar h_m-\widehat{\operatorname{Cov}}(h,c)^\top
\widehat\Sigma_c^{-1}\bar c_m .
\label{eq:greg-identity}
\end{equation}
For a scalar target, TCRF is therefore exactly a same-sample regression-calibration or control-variate estimator in finite samples, and we do not claim a new regression adjustment. The methodological contribution is to construct the calibration functions from downstream learning influence and then to use the resulting common frequency weights as the random-feature representation. For any fixed $\beta\in\mathbb R^r$, define the residual
$e(Z)=u(Z)-\beta^\top c(Z)$. Exact balance gives
\[
\hat\theta_{\mathrm{TCRF}}-\theta
=\frac1m\sum_{j=1}^m w_je(Z_j),
\]
so that the part of the Monte Carlo fluctuation lying in the calibration space is removed exactly on the observed frequency sample.

To describe the remaining fluctuation, we suppose throughout that $u$ and $c$ are square integrable under $q_0$ and that $\Sigma_c=\mathbb{E}_{q_0}[c(Z)c(Z)^\top]$ is positive definite; these conditions hold for the bounded calibration functions of kernel ridge regression and, under the Gaussian spectral law, for those of the Stein discrepancy. Let $\mathcal C=\operatorname{span}\{c_1,\ldots,c_r\}$, and let $\Pi_{\mathcal C}$ denote the orthogonal projection onto $\mathcal C$ in $L_2(q_0)$. By the normal equations, $\Pi_{\mathcal C}u=\beta^\top c$ with $\beta=\Sigma_c^{-1}\mathbb{E}_{q_0}[c(Z)u(Z)]$, and the corresponding residual satisfies $\mathbb{E}_{q_0}[c(Z)e(Z)]=0$.

\begin{proposition}[First-order variance projection]
\label{prop:task-calibration}
As $m\to\infty$,
\[
\sqrt m(\hat\theta_{\mathrm{TCRF}}-\theta)
=\frac1{\sqrt m}\sum_{j=1}^m
\bigl(u(Z_j)-\Pi_{\mathcal C}u(Z_j)\bigr)+o_p(1),
\]
and the asymptotic variance under $\sqrt m$ scaling is
\[
\operatorname{avar}(\hat\theta_{\mathrm{TCRF}})
=\left\|(I-\Pi_{\mathcal C})u\right\|_{L_2(q_0)}^2.
\]
\end{proposition}

\begin{proof}
Put $\hat\beta=\widehat\Sigma_c^{-1}\widehat{\operatorname{Cov}}(h,c)$.
By square integrability and the Cauchy--Schwarz inequality,
\[
\mathbb{E}_{q_0}[\|c(Z)u(Z)\|]
\le\bigl(\mathbb{E}_{q_0}[\|c(Z)\|^2]\bigr)^{1/2}
\bigl(\mathbb{E}_{q_0}[u(Z)^2]\bigr)^{1/2}<\infty,
\]
so the laws of large numbers give
$\widehat\Sigma_c\xrightarrow{p}\Sigma_c$ and
$\widehat{\operatorname{Cov}}(h,c)\xrightarrow{p}\mathbb{E}_{q_0}[c(Z)u(Z)]$.
Since $\Sigma_c$ is positive definite, $\widehat\Sigma_c$ is nonsingular
with probability tending to one and $\hat\beta\xrightarrow{p}\beta$.
Furthermore, $\mathbb{E}_{q_0}[c(Z)]=0$ and the central limit theorem yield
$\sqrt m\,\bar c_m=O_p(1)$. The identity \eqref{eq:greg-identity} now gives
\[
\sqrt m(\hat\theta_{\mathrm{TCRF}}-\theta)
=\frac1{\sqrt m}\sum_{j=1}^m
\bigl(u(Z_j)-\beta^\top c(Z_j)\bigr)-(\hat\beta-\beta)^\top\sqrt m\,\bar c_m,
\]
whose second term is $o_p(1)$. Since $\beta^\top c=\Pi_{\mathcal C}u$ and
$u-\beta^\top c$ is centred and square integrable, the i.i.d. central
limit theorem proves the asymptotic expansion and its variance formula.
\end{proof}

Since $\mathcal C$ is finite dimensional and hence closed in $L_2(q_0)$,
this asymptotic variance vanishes if and only if $u\in\mathcal C$.
If $u=a^\top c$ holds $q_0$-almost surely, the exact balance
\eqref{eq:exact-balance} further gives
$\hat\theta_{\mathrm{TCRF}}=\theta$ on every sample for which
$\widehat\Sigma_c$ is nonsingular. Thus complete containment removes
the fluctuation in finite samples, while approximate containment leaves
the squared norm of its orthogonal residual as the first-order variance.
The closed-form weights need not be nonnegative on every finite sample. For fixed $r$, if $\max_{j}\|c(Z_j)\|=o_p(\sqrt m)$ and $\Sigma_c$ is positive definite, then $\max_j|w_j-1|=o_p(1)$ and the probability that all weights are positive tends to one. In kernel ridge regression the calibration functions are bounded combinations of cosine features. In the Stein discrepancy they have quadratic growth in the frequency, but Gaussian spectral tails give a maximum of logarithmic order in fixed dimension, so the required control holds there as well.

\paragraph{Preservation of the weighted feature representation.}
The identity \eqref{eq:greg-identity} makes the relation to control variates exact rather than merely first order, and the algorithmic distinction lies in how the weights are used. The controls are derived from the downstream learning influence, and the resulting weights are shared by every occurrence of a sampled frequency in every kernel entry. If $w_j\ge0$ for all $j$, then, with the feature block $\phi_Z$ of Section~\ref{sec:framework},
\begin{equation}
\phi_w(x)
=
\frac1{\sqrt m}
\left(
\sqrt{w_1}\phi_{Z_1}(x)^\top,\ldots,
\sqrt{w_m}\phi_{Z_m}(x)^\top
\right)^\top
\label{eq:weighted-feature-map}
\end{equation}
is a valid feature map, and its inner product is the weighted random-feature kernel, namely the average of $w_j\cos(Z_j^{\top}(x-x')/\sigma)$ over the sampled frequencies. Hence nonnegative common weights preserve positive semidefiniteness, and the normalisation in \eqref{eq:exact-balance} preserves the unit diagonal. For the Stein discrepancy every frequency contribution is nonnegative, so the weighted squared discrepancy also remains nonnegative. A generic signed scalar or entrywise control-variate correction need not preserve either representation, which is why a weight-based formulation is preferred here even though the first-order projection is shared with control variates.

\paragraph{Closed-form rank selection.}
Because the weights are invariant to an invertible linear reparameterisation
of the calibration functions, a conditioning diagnostic should not
depend on the arbitrary scale of each direction. For a candidate rank
$r\in\{1,\ldots,r_{\max}\}$, let $c_{1:r}=(c_1,\ldots,c_r)^\top$,
let $\widehat\Sigma_{c,r}$ denote its centred empirical covariance
on the $m$ main frequencies, and let $\operatorname{diag}(\widehat\Sigma_{c,r})$
be its diagonal part. We consider only candidates
for which $\widehat\Sigma_{c,r}$ is nonsingular, and define the empirical
correlation matrix
\[
\widehat R_{c,r}=\operatorname{diag}(\widehat\Sigma_{c,r})^{-1/2}\,
\widehat\Sigma_{c,r}\,\operatorname{diag}(\widehat\Sigma_{c,r})^{-1/2}.
\]
It has unit diagonal, but it need
not be the identity because the calibration directions may be correlated.
Our scale-free conditioning score is
\[
\chi_r=\frac{\lambda_{\min}(\widehat R_{c,r})}
{\lambda_{\max}(\widehat R_{c,r})}.
\]
For each such candidate, let $w_{j,r}$ be the closed-form weight
\eqref{eq:closed-form-weight} computed with $c_{1:r}$ in place of $c$,
for $j=1,\ldots,m$, and put $w_{j,0}=1$. We select the rank by
\begin{equation}
\hat r=\max\left\{
 r\in\{1,\ldots,r_{\max}\}:\;\chi_r\ge\kappa_{\mathrm{cond}},\quad
 \min_{1\le j\le m}w_{j,r}\ge\varepsilon
\right\}.
\label{eq:rank-rule}
\end{equation}
If no positive rank satisfies these conditions, we set $\hat r=0$;
throughout, $r_{\max}<m$.
The fixed-rank theory above and the adaptive choice of $\hat r$ are kept conceptually separate.

In the population, the residual variance decreases monotonically in the rank. In finite samples, by contrast, increasing $r$ makes the estimation of $\widehat\Sigma_{c,r}$ less stable and the weights more variable. For the Gaussian benchmark of classical control variates the mean squared error including coefficient estimation carries the loss factor $(m-2)/(m-r-2)$, so that adding a further direction that removes population variance $\Delta_{r+1}$ from a residual variance $V_r$ is worthwhile only when $\Delta_{r+1}>V_r/(m-r-2)$, a criterion implied by the classical loss factor discussed by \citet{nelson1990}. We do not adopt this approximation as the rank-selection rule; instead, \eqref{eq:rank-rule} screens for positivity and conditioning. For each candidate rank only $\bar c_m$ and an $r\times r$ covariance matrix are needed, and no nonlinear optimisation is involved.

\paragraph{Positive exponential-tilting calibration.}
As an optional implementation extension of the closed-form construction,
motivated by classical exponential-tilting calibration
\citep{kim2010}, we consider the KL calibration problem
\begin{equation}
\min_{w_1,\ldots,w_m>0}\frac1m\sum_{j=1}^m
(w_j\log w_j-w_j+1)
\quad\text{subject to}\quad
\frac1m\sum_{j=1}^m w_j=1,\quad
\frac1m\sum_{j=1}^m w_jc(Z_j)=0.
\label{eq:et-primal}
\end{equation}
Writing $c_j=c(Z_j)$, the constrained problem
\eqref{eq:et-primal} has, when a finite dual solution exists,
normalised exponential-tilting (ET) weights
\begin{equation}
w_j^{\mathrm{ET}}(\xi)=
\frac{\exp(-\xi^\top c_j)}
{m^{-1}\sum_{\ell=1}^m\exp(-\xi^\top c_\ell)},
\quad
\frac1m\sum_{j=1}^m w_j^{\mathrm{ET}}(\xi)c_j=0.
\label{eq:et-weights}
\end{equation}
The dual variable $\xi$ is distinct from the Gaussian sampling
parameter $\eta$ in Section~\ref{sec:framework}.
The ET weights are strictly positive and satisfy exact empirical balance
up to numerical tolerance. Positivity itself does not ensure existence:
when $\widehat\Sigma_c$ is nonsingular, a finite dual solution exists if and only if
the origin lies in the interior of the convex hull of the
observed control vectors. Our implementation
checks dual convergence and the balance residual for each candidate
rank, and drops infeasible candidates.

The relation to the closed-form formula is local. For a fixed-dimensional
control vector with zero population mean, nonsingular $\Sigma_c$, and
sufficient moment and regularity conditions for the Taylor expansion,
the ET multiplier obeys
\[
\hat\xi=\widehat\Sigma_c^{-1}\bar c_m+O_p(m^{-1}),
\]
and the exponential weights have the first-order expansion
\[
w_j^{\mathrm{ET}}
=1-(c_j-\bar c_m)^\top\widehat\Sigma_c^{-1}\bar c_m
+O_p(m^{-1})
\]
for each fixed index $j$. Thus the closed-form weights are the
\emph{linearised} ET calibration weights; \eqref{eq:greg-identity}
is an exact finite-sample identity for the closed-form case, not for ET.
For fixed rank and exactly centred controls, both constructions have
the same first-order projection variance under the corresponding
regularity conditions. The second-order KRR limit in
Theorem~\ref{prop:krr-quadratic-limit}, however, is proved for the
closed-form estimator and is not asserted automatically for ET,
growing selected ranks, or approximately centred controls.

\subsection{Kernel ridge regression}
\label{sec:krr-main}

Throughout this subsection, the training data, $n$, $\lambda>0$ and
test point $x$ are fixed; all randomness comes from sampled frequencies.
Using the same closed-form weights for the training Gram matrix and the
test-to-training kernel vector, put
\[
\widehat K_w=\frac1m\sum_{j=1}^m w_jA_{Z_j},
\quad
\hat k_{x,w}=\frac1m\sum_{j=1}^m w_ja_{Z_j}(x),
\]
which, for nonnegative weights, are the Gram matrix and kernel vector of the
feature map \eqref{eq:weighted-feature-map}, and define the calibrated predictor by
\[
\hat f_{\mathrm{TCRF}}(x)=
\hat k_{x,w}^{\top}(\widehat K_w+n\lambda I_n)^{-1}y.
\]
By Proposition~\ref{prop:task-calibration}, the first-order variance
of this predictor associated with a calibration space $\mathcal C$ is
\[
V_x(\mathcal C)
=\left\|(I-\Pi_{\mathcal C})b_x\right\|_{L_2(q_0)}^2.
\]
When $b_x\in\mathcal C$, the weighted first-order term vanishes
exactly and $\sqrt m(\hat f_{\mathrm{TCRF}}(x)-f_K(x))\to_p0$.
The predictor is nonlinear in the kernel matrix, however, so a
second-order error remains. The next theorem identifies its limit.
The centred frequency vector and its projection residual are
\[
H_x(Z)=
\begin{pmatrix}
\operatorname{vec}(A_Z-K)\\
a_Z(x)-k_x
\end{pmatrix},
\quad
H_x^\perp(Z)=H_x(Z)-\mathbb{E}(H_x(Z)c(Z)^\top)\Sigma_c^{-1}c(Z).
\]
Let $(\mathcal U,\mathcal V)$ be a centred jointly Gaussian
matrix--vector pair with
\[
\begin{pmatrix}\operatorname{vec}(\mathcal U)\\\mathcal V\end{pmatrix}
\sim N(0,\Omega_x),\quad
\Omega_x=\mathbb{E}\bigl[H_x^\perp(Z)H_x^\perp(Z)^\top\bigr].
\]

\begin{theorem}[Second-order limit after exact cancellation]
\label{prop:krr-quadratic-limit}
Suppose that $b_x\in\mathcal C$. Then, as $m\to\infty$,
\begin{equation}
m(\hat f_{\mathrm{TCRF}}(x)-f_K(x))
\xrightarrow{d}
Q_x=-(\mathcal V-\mathcal U Rk_x)^\top R\mathcal U\alpha.
\label{eq:krr-quadratic-limit}
\end{equation}
\end{theorem}

\begin{proof}
Put $E_m=\widehat K_w-K$, $e_m=\hat k_{x,w}-k_x$ and
$d_m=e_m-E_mRk_x$.
The fixed-data setup makes $K$, $k_x$, $R$ and $\alpha$ fixed,
with $R$ nonsingular because $\lambda>0$.
The entries of $A_Z$ and $a_Z(x)$ are bounded for fixed training and
test points, so Proposition~\ref{prop:task-calibration} applies to each
component of $H_x$, and $\widehat\Sigma_c$ is invertible with
probability tending to one. Because the dimension of $H_x$ is fixed,
the multivariate central limit theorem yields the joint convergence
\begin{equation}
\left(\sqrt m E_m,\sqrt m e_m\right)
\xrightarrow{d}(\mathcal U,\mathcal V).
\label{eq:krr-joint-clt}
\end{equation}
Consequently $E_m=O_p(m^{-1/2})$ and $e_m=O_p(m^{-1/2})$.

Define $\hat\alpha_m=(K+E_m+n\lambda I_n)^{-1}y$.
Since $K+n\lambda I_n$ is nonsingular and $E_m=o_p(1)$, the
inverse defining $\hat\alpha_m$ exists with probability tending
to one and $\hat\alpha_m\xrightarrow{p}\alpha$.
The resolvent identity gives
\[
\hat\alpha_m=\alpha-RE_m\hat\alpha_m.
\]
Using the definitions of $d_m$ and $\hat f_{\mathrm{TCRF}}$,
we obtain the exact identity
\begin{equation}
\hat f_{\mathrm{TCRF}}(x)-f_K(x)
=d_m^{\top}\hat\alpha_m
=d_m^{\top}\alpha+d_m^{\top}(\hat\alpha_m-\alpha)
=d_m^{\top}\alpha-d_m^{\top}RE_m\hat\alpha_m.
\label{eq:krr-exact-error}
\end{equation}
Because $b_x\in\mathcal C$, we have $b_x=\beta_x^\top c$ as an equality in
$L_2(q_0)$ for a fixed vector $\beta_x$. Thus, for the independently
sampled frequencies, this equality holds almost surely. The common
weights and exact calibration then give
\[
d_m^\top\alpha
=\frac1m\sum_{j=1}^m w_j b_x(Z_j)
=\beta_x^\top\frac1m\sum_{j=1}^m w_jc(Z_j)=0.
\]
Consequently the first term on the right of
\eqref{eq:krr-exact-error} vanishes exactly.
Finally, multiply the remaining term by $m$. The joint limit
\eqref{eq:krr-joint-clt}, $\hat\alpha_m\xrightarrow{p}\alpha$
and the continuous mapping theorem yield \eqref{eq:krr-quadratic-limit}.
\end{proof}

The prediction error is thus of order $m^{-1}$, unless the limit is degenerate. The limit $Q_x$ is a quadratic form in correlated Gaussian kernel errors, distributed as a weighted sum of independent chi-square variables with one degree of freedom; the weights can have either sign. Consequently its expectation need not vanish, and bias and standard deviation can be of the same order. Weak convergence alone does not establish convergence of finite-sample bias or mean squared error without suitable uniform integrability. When $Q_x=0$ almost surely, still faster convergence or even exact finite-sample agreement may occur.

Theorem~\ref{prop:krr-quadratic-limit} concerns frequency-sampling randomness conditional on the fixed training data; it does not involve $n\to\infty$ or $\lambda\to0$. It applies to independent original frequencies with one-stage exact calibration. A hybrid using prior moment matching or an estimated calibration space requires separate asymptotic justification. In particular, when $V_x(\mathcal C)>0$ for a fixed space, the usual $m^{-1/2}$ fluctuation remains even if this variance is small. For spaces varying with $m$, the residual first-order norm must be of order $m^{-1/2}$ to balance the quadratic contribution, with appropriate uniform control of pilot estimation and conditioning. The experiments below assess near-zero variance; they do not establish this second-order limit for adaptive procedures.

For representative points $x_1^*,\ldots,x_J^*$, let
\[
B(Z)=\bigl(b_{x_1^*}(Z),\ldots,b_{x_J^*}(Z)\bigr)^\top,
\quad \Sigma_B=\mathbb{E}[B(Z)B(Z)^\top].
\]
Let $\nu_1\geq\cdots\geq\nu_J\geq0$ denote its eigenvalues and
$v_1,\ldots,v_J$ corresponding eigenvectors, and set
\[
c_k(Z)=v_k^{\top}B(Z),
\quad
\mathcal C_r=\operatorname{span}\{c_1,\ldots,c_r\}.
\]
In implementation only directions corresponding to positive eigenvalues are used. This choice has a clear optimality property with respect to near-zero variance: for $1\le r\le\operatorname{rank}(\Sigma_B)$ with $\nu_r>\nu_{r+1}$, where $\nu_{J+1}=0$, the space $\mathcal C_r$ is the best $r$-dimensional calibration space for the representative influence family.

\begin{proposition}[Optimal rank-$r$ calibration space]
\label{prop:principal-calibration}
The subspace $\mathcal C_r$ is the unique solution of
\[
\min_{\mathcal C\subset L_2(q_0),\ \dim(\mathcal C)=r}
\sum_{\ell=1}^J
\left\|(I-\Pi_{\mathcal C})b_{x_\ell^*}\right\|_{L_2(q_0)}^2,
\]
and the minimum equals $\sum_{k=r+1}^J\nu_k$.
\end{proposition}

\begin{proof}
The influence functions are bounded, so $\Sigma_B$ is finite, symmetric
and positive semidefinite. For $\nu_k>0$ put
$\varphi_k=\nu_k^{-1/2}v_k^\top B$. Since
$\mathbb{E}_{q_0}[(v_k^\top B(Z))(v_{k'}^\top B(Z))]=v_k^\top\Sigma_Bv_{k'}=\nu_k\delta_{kk'}$,
these functions are orthonormal in $L_2(q_0)$ and span the linear span of
$b_{x_1^*},\ldots,b_{x_J^*}$; moreover
$\sum_{\ell=1}^J\langle b_{x_\ell^*},\varphi_k\rangle\langle b_{x_\ell^*},\varphi_{k'}\rangle=\nu_k\delta_{kk'}$.
Let $\mathcal C$ be an $r$-dimensional subspace with orthogonal projection
$P$ and orthonormal basis $e_1,\ldots,e_r$, and put
$t_k=\|P\varphi_k\|^2\in[0,1]$. By Bessel's inequality,
$\sum_kt_k=\sum_{j=1}^r\sum_k\langle e_j,\varphi_k\rangle^2\le r$.
Expanding each $b_{x_\ell^*}$ in the functions $\varphi_k$ gives
$\sum_{\ell=1}^J\|Pb_{x_\ell^*}\|^2=\sum_k\nu_kt_k$, so that the
objective equals $\sum_{k=1}^J\nu_k-\sum_k\nu_kt_k$. Because
$\nu_k\le\nu_{r+1}$ for $k>r$ and $\sum_{k>r}t_k\le r-\sum_{k\le r}t_k$,
\[
\sum_k\nu_kt_k
\le\sum_{k=1}^r\nu_k-\sum_{k=1}^r(\nu_k-\nu_{r+1})(1-t_k)
\le\sum_{k=1}^r\nu_k .
\]
Hence the objective is at least $\sum_{k>r}\nu_k$, with equality for
$\mathcal C_r=\operatorname{span}\{\varphi_1,\ldots,\varphi_r\}$, for which
$t_k=1$ when $k\le r$ and $t_k=0$ otherwise. Conversely, if $\mathcal C$
attains the minimum, both inequalities are equalities. Since
$\nu_k-\nu_{r+1}\ge\nu_r-\nu_{r+1}>0$ for $k\le r$, this forces $t_k=1$,
that is, $\varphi_k\in\mathcal C$ for $k\le r$, and $\dim\mathcal C=r$
gives $\mathcal C=\mathcal C_r$.
\end{proof}

Since the total variance of the representative influence family is
$\operatorname{tr}(\Sigma_B)=\sum_{k=1}^J\nu_k>0$, the proportion
of first-order variance removed by $\mathcal C_r$ is
\[
\frac{\sum_{k=1}^r\nu_k}{\sum_{k=1}^J\nu_k}.
\]
When the eigenvalue tail is small, a few calibration directions can capture most of the task-relevant variation. An influence matrix evaluated on $M_0$ independent pilot frequencies estimates these principal directions by singular value decomposition. The resulting method approximates a task-specific zero-variance space, rather than reducing arbitrary kernel-entry errors. Its prediction influence is response-adaptive through $\alpha=Ry$; this paper focuses on approximating the fitted prediction, whereas response-independent spectral approximation addresses a different objective.

Task-derived directions can overlap with Gaussian covariance-score
directions, as the Hermite calculation in Appendix~\ref{supp:mm}
shows. This suggests augmenting the task directions by covariance-score
directions in a single calibration step, with an effective rank that
depends on their overlap. We also examine a two-stage hybrid that
centres and whitens the frequencies before closed-form calibration.
Its transformed frequencies are dependent, so the independent-frequency
projection theorem does not directly describe that hybrid.

Computing the exact influence requires the exact matrices $K$ and $R$ and the exact coefficient vector $\alpha$, which defeats the purpose when the goal is to approximate large-scale kernel ridge regression. We therefore select $n_L\ll n$ landmarks from the training sample and solve a small exact problem. Writing $K_L$ for the landmark Gram matrix, $y_L$ for the landmark responses, $R_L=(K_L+n_L\lambda I_{n_L})^{-1}$, $\alpha_L=R_Ly_L$, $k_{x,L}$ for the exact kernel vector between a test point and the landmarks, and $A_{Z,L}$ and $a_{Z,L}(x)$ for the landmark analogues of $A_Z$ and $a_Z(x)$, the landmark influence is
\[
b_x^{(L)}(Z)
=
\left(a_{Z,L}(x)-k_{x,L}\right)^{\top}\alpha_L
-
k_{x,L}^{\top}R_L(A_{Z,L}-K_L)\alpha_L ,
\]
whose expectation under the spectral distribution is zero, so valid calibration directions are produced from small exact kernel computations alone. We learn the principal space of this family on pilot frequencies and calibrate the main frequency sample with it, using the full exact solution only as the benchmark. Complete zero variance with respect to the full task influence is not attained in general, but if the landmark influence space approximates the dominant directions of the full influence space then the residual norm is small and near-zero variance can be expected. The landmark bias is accordingly to be interpreted not as a generic nuisance approximation error but as the approximation error of a zero-variance space, measured by the residual criterion of Proposition~\ref{prop:principal-calibration} evaluated at the full influence.
The large-sample Nystr\"om--Taylor implementation adds a distinct issue:
approximating the full Gram matrix in the centring term can introduce a
nonzero spectral mean even when the empirical calibration equations are
solved exactly. The identity and an independent centring diagnostic are
given in Appendix~\ref{sec:supp-large-krr}.

\subsection{Kernel Stein discrepancy}

For the Stein discrepancy the computation of Section~\ref{sec:framework} is exactly the Monte Carlo target of Proposition~\ref{prop:task-calibration}, with $u=u_D$. The conditional first-order variance associated with a calibration space $\mathcal C$ is
\[
V_D(\mathcal C)
=
\left\|(I-\Pi_{\mathcal C})u_D\right\|_{L_2(q_0)}^2,
\quad
\rho_D(\mathcal C)
=
\frac{V_D(\mathcal C)}{\operatorname{Var}_\omega(G_D(\omega))},
\]
and if a fixed $\mathcal C$ contains $u_D$ then the weighted discrepancy agrees with $\theta_D$ exactly on any sample for which $\widehat\Sigma_c$ is nonsingular. No quadratic remainder of the kind appearing in Theorem~\ref{prop:krr-quadratic-limit} survives, because $G_D$ is itself the frequency integrand. This statement concerns the integration error for a fixed dataset and must be distinguished from the null distribution of a test in which $n$ grows. The residual ratio $\rho_D$ measures the degree of near-zero variance, and a value close to zero indicates that a computation close to the exact empirical discrepancy is attainable with few frequencies. Using the centred integrand itself as a calibration function would require the exact discrepancy and would defeat the purpose, so the question is how to obtain a cheap surrogate. Random subsets of the data provide one. For a subset $S$ of size $s$, define its frequency integrand by
\[
G_S(\omega)=\left\|\frac1s\sum_{i\in S}\psi_\omega(X_i)\right\|^2,
\]
and define the centred control, isotropic direction and full-data
 diagonal contribution by
\[
\begin{aligned}
c_S(\omega)&=G_S(\omega)-\mathbb{E}_\omega[G_S(\omega)],\\
q(\omega)&=\|\omega\|^2-\mathbb{E}_\omega\|\omega\|^2,\\
A_D(\omega)&=\frac1n\sum_{i=1}^n\|\psi_\omega(X_i)\|^2.
\end{aligned}
\]
Let $S$ be a simple random sample of size $1\le s\le n$ from the fixed dataset, with $n\ge2$.
The inclusion indicators $I_i=\mathbf1\{i\in S\}$ satisfy
\[
\mathbb{E}_S[I_i]=\frac{s}{n},\qquad
\mathbb{E}_S[I_iI_j]=\frac{s(s-1)}{n(n-1)}\quad (i\ne j).
\]
Expand $G_S=s^{-2}\sum_{i,j}I_iI_j\psi_\omega(X_i)^\top\psi_\omega(X_j)$,
separate diagonal from off-diagonal terms, and use the corresponding
full-data identity for $G_D$. We obtain for every frequency
\[
\mathbb{E}_S(G_S(\omega))
=a_{n,s}G_D(\omega)+b_{n,s}A_D(\omega),
\quad
a_{n,s}=\frac{n(s-1)}{s(n-1)},
\quad
b_{n,s}=\frac{n-s}{s(n-1)},
\]
with $a_{n,s}+b_{n,s}=1$. For the Langevin Stein feature,
$A_D(\omega)=n^{-1}\sum_i\|s_p(X_i)\|^2+\|\omega\|^2$, so centring with
respect to the spectral law yields the subset identity
\begin{equation}
\mathbb{E}_S(c_S(\omega))
=a_{n,s}u_D(\omega)+b_{n,s}q(\omega).
\label{eq:subset-identity}
\end{equation}
A subset average is therefore a noisy surrogate for the full-data zero-variance direction, biased only along the isotropic direction $q$. If $\bar c$ is an average of subset functions whose expectation is $a_{n,s}u_D+b_{n,s}q$, the two-dimensional space $\mathcal C=\operatorname{span}\{\bar c,q\}$ contains $u_D$ up to the subset noise. Given $D$ and the subsets, $\bar c$ is a fixed function of the frequency, so Proposition~\ref{prop:task-calibration} applies conditionally and no pilot frequencies are required. The function $q$ is proportional to the isotropic Gaussian scale score, so the correction needed to pass from a subset average to the full integrand appears automatically as one of the directions contained in moment matching; this does not mean that the whole covariance-score effect degenerates to a single direction, only that one additional direction suffices for the subset-to-full correction. The integrand and the subset functions have quadratic growth in the frequency, since the empirical characteristic function of a finite dataset does not decay uniformly. Because the closed-form construction introduces no global sampler density, what is required is only stability of second moments and of the sample covariance in $L_2(q_0)$, which holds under the Gaussian spectral law. 

There is, however, a computational point that the identity alone does not settle. The centring constant $\mathbb{E}_\omega(G_S(\omega))$ is the exact V-statistic on the subset and requires $s^2$ evaluations of the Stein kernel. With $L$ independent subsets the pilot needs $Ls^2$ pair evaluations, and this can exceed the $n^2$ evaluations of the exact statistic; for instance $n=500$, $s=200$ and $L=30$ give $1.2\times10^6$ against $2.5\times10^5$. The design of the subsets is therefore part of the computational problem.

To analyse the subset noise, fix $\omega$ and consider for $i\neq j$ the centred pair function
\[
\kappa_{ij}(\omega)=\psi_\omega(X_i)^\top\psi_\omega(X_j)-h_p(X_i,X_j),
\]
which satisfies $\mathbb{E}_\omega(\kappa_{ij}(\omega))=0$. Since the diagonal terms centre to $q$, one has $c_S=s^{-1}q+s^{-2}\sum_{i\neq j\in S}\kappa_{ij}$ and $u_D=n^{-1}q+n^{-2}\sum_{i\neq j}\kappa_{ij}$. For $n\ge3$ the symmetric array $(\kappa_{ij})_{i\neq j}$ has a unique finite-population Hoeffding decomposition
\[
\kappa_{ij}=\mu+\alpha_i+\alpha_j+\gamma_{ij},
\quad
\sum_{i=1}^n\alpha_i=0,
\quad
\sum_{j\neq i}\gamma_{ij}=0\ \text{ for every } i,
\]
in which $\mu$ is the average of the pair functions, the linear terms $\alpha_i$ are the finite-population analogues of the H\'ajek projection, and $\gamma_{ij}$ is the doubly centred, or degenerate, part (Appendix Lemma~\ref{lem:hoeffding}). All components are functions of $\omega$, and we write
\[
\bar A=\frac1n\sum_{i=1}^n\|\alpha_i\|_{L_2(q_0)}^2,
\quad
\bar\Gamma=\frac1{n(n-1)}\sum_{i\neq j}\|\gamma_{ij}\|_{L_2(q_0)}^2 ,
\]
which are finite under the Gaussian spectral law.
We compare two pilot designs with $L$ subsets of size $s$, where $n\ge4$, $2\le s\le n$ and $N=Ls\le n$. In the independent design, $S_1,\ldots,S_L$ are independent simple random samples. In the disjoint-block design, the first $N$ entries of a uniformly random permutation of $\{1,\ldots,n\}$ are cut into $L$ consecutive blocks, so that the blocks are disjoint and, when $N=n$, form a uniformly random partition of the data. In both designs we write $\bar c=L^{-1}\sum_{l=1}^Lc_{S_l}$.

\begin{theorem}[Disjoint-block variance decomposition]
\label{prop:ksd-blocks}
For a fixed dataset $D$, the disjoint-block design satisfies
\begin{equation}
\mathbb{E}\|\bar c-\mathbb{E}\bar c\|_{L_2(q_0)}^2
=
\frac{4(s-1)^2(n-N)}{Ns^2(n-1)}\,\bar A
+
\frac{2(s-1)\big((n-s)(n-s-1)+(N-s)(s-1)\big)}{Ns^2(n-2)(n-3)}\,\bar\Gamma,
\label{eq:block-variance}
\end{equation}
whereas the independent design satisfies
\begin{equation}
\mathbb{E}\|\bar c-\mathbb{E}\bar c\|_{L_2(q_0)}^2
=
\frac{4(s-1)^2(n-s)}{Ns^2(n-1)}\,\bar A
+
\frac{2(s-1)(n-s)(n-s-1)}{Ns^2(n-2)(n-3)}\,\bar\Gamma.
\label{eq:independent-variance}
\end{equation}
\end{theorem}

\begin{proof}
The finite-population Hoeffding decomposition separates the subset
fluctuation into linear and degenerate terms. In a full random
partition every observation has exactly $s-1$ blockmates, so the
linear term cancels identically. Computing the remaining pairwise
selection covariances gives the displayed variance formulas; the full
enumeration appears in Appendix~\ref{supp:ksd}.
\end{proof}

Because each subset is marginally a simple random sample of size $s$,
\eqref{eq:subset-identity} shows that both designs have the same mean function
$\mathbb{E}\bar c(\omega)=a_{n,s}u_D(\omega)+b_{n,s}q(\omega)$.
Consequently, for $s\ge2$ and
$\mathcal C=\operatorname{span}\{\bar c,q\}$,
\[
\mathbb{E}\bigl[\operatorname{dist}^2(u_D,\mathcal C)\bigr]
\le a_{n,s}^{-2}\,
\mathbb{E}\|\bar c-\mathbb{E}\bar c\|_{L_2(q_0)}^2.
\]
When $N=n$ the blocks form a random partition, and the first term of \eqref{eq:block-variance} vanishes: the balancing mechanism that underlies the calibration weights here acts on the data. The dependence between blocks is therefore not a nuisance to be controlled but the source of the improvement. Comparing \eqref{eq:block-variance} with \eqref{eq:independent-variance} at equal $N$ and $s$, the independent design minus the disjoint design equals
\[
\frac{2(s-1)^2(N-s)}{Ns^2}\left(\frac{2\bar A}{n-1}-\frac{\bar\Gamma}{(n-2)(n-3)}\right).
\]
Disjoint blocks are thus preferable whenever $\bar A\ge(n-1)\bar\Gamma/(2(n-2)(n-3))$, a condition that does not involve $s$ or $N$, and for $s<n-1$ the price paid otherwise is a degenerate term larger by a factor of at most $1+(s-1)/(n-s-1)$. The same formula covers a number of observations that is not a multiple of $s$: taking $L$ as the integer part of $n/s$ leaves fewer than $s$ observations outside the blocks, and the linear term then carries the factor $n-N<s$ in place of $n-s$. Averaging $P$ independent partitions divides \eqref{eq:block-variance} by $P$.

The two displays also quantify the subset-noise contribution to an upper
bound on the residual ratio. The number of pair evaluations needed for the
exact centring constants is $Ns$. For a full partition with $s\ll n$,
Theorem~\ref{prop:ksd-blocks} bounds the squared subset noise approximately
by $2\bar\Gamma/(ns)$, and division by
$a_{n,s}^2\operatorname{Var}_\omega(G_D(\omega))$ gives a corresponding
relative-error upper bound. Under a fixed alternative with a spectral
signal bounded away from zero and bounded projected pair moments, the
resulting upper bound is of order $1/(ns)$; a fixed subset size therefore
gives a subquadratic centring cost as $n$ increases. Under the null
hypothesis, the usual degenerate-U-statistic scaling makes the full-data
spectral fluctuation small, so the subset-noise upper bound need not
provide a useful relative-error guarantee when $s\ll n$. This limitation
of the bound does not imply that a pilot with quadratic cost is necessary.
The bound uses the fixed
coefficient $a_{n,s}^{-1}$, whereas calibration projects onto $\mathcal C$
with optimal coefficients; Appendix Proposition~\ref{prop:exact-residual}
gives the exact residual, which depends on the alignment of subset noise
with the signal.

The computational cost of the full-partition pilot is as follows. The centring constants require $ns$ evaluations of the Stein kernel, at cost of order $nsd$. The values of the subset functions at the main frequencies are obtained from block sums of the vectors $\psi_{\omega_j}(X_i)$, which are computed anyway for $G_D(\omega_j)$, so the calibration functions add only order $mLd$ operations, and the closed-form weights for a two-dimensional space add order $m$. The calibrated random-feature discrepancy therefore costs order $(m+s)nd$, compared with order $n^2d$ for the exact statistic, and it requires neither pilot frequencies nor additional feature evaluations. The independent design with the same budget has the same degenerate term, up to the factor above, and an additional linear term of approximate size $4\bar A/n$, which does not decrease as $s$ grows; this is the term that the partition removes.

\section{Numerical experiment}
\label{sec:numerical}

We evaluate the cancellation mechanism in paired Monte Carlo
experiments. All comparisons within a replication use the same training
and test samples and nested frequency sequences; independent
replications use fixed, documented random seeds. For the kernel ridge
regression experiments we use $d=5$,
\[
X_i\sim N(0,I_5),
\quad
y_i=f_0(X_i)+\varepsilon_i,
\quad
\varepsilon_i\sim N(0,0.5^2),
\]
\[
f_0(x)=\sin(x_1)+0.6x_2x_3+0.35(x_4^2-1),
\]
with $n=300$, a test set of size $1000$, and Gaussian-kernel bandwidth equal to the median pairwise distance in the training sample. Our primary predictive evaluation selects the ridge parameter by five-fold cross-validation within each replication, over 26 logarithmically spaced values in $[10^{-6},10^{-1}]$ under the convention $(K+n\lambda I_n)^{-1}$. We also report the fixed $\lambda=0.02$ setting to isolate the calibration mechanism. For a controlled comparison of frequency approximations, cross-validation uses the \emph{exact} training Gram matrix; it is a reference-tuning experiment, not an inexpensive RF-based tuning algorithm for very large $n$. Every method within a replication uses the same fitted ridge parameter and paired training, test, and frequency draws; because $n=300$, the random-feature predictors are computed by solving the $n\times n$ system directly. Closed-form weights use \eqref{eq:rank-rule} with $\kappa_{\mathrm{cond}}=10^{-4}$ and $\varepsilon=0.01$; ET uses \eqref{eq:et-rank-rule} and the same conditioning screen. Monte Carlo errors and sensitivity results are reported in Appendix~\ref{supp:numerical}, together with a separate large-sample closed-form benchmark (Appendix~\ref{sec:supp-large-krr}).

\subsection{Kernel ridge regression}

The closed-form method uses \eqref{eq:rank-rule}. For ET, a converged
positive solution replaces the minimum-weight screen, subject to the
same conditioning and feasibility checks. The full-KRR experiments
select the feasible rank (including $r=0$) that minimises the pilot proxy
\begin{equation}
\hat r_{\mathrm{ET}}
=\arg\min_{r\in\mathcal R_{\mathrm{feas}}}
\widehat V_r
\left(\frac1m\sum_{j=1}^m(w_{j,r}^{\mathrm{ET}})^2\right).
\label{eq:et-rank-rule}
\end{equation}
Here $w_{j,r}^{\mathrm{ET}}$ are the weights \eqref{eq:et-weights} computed
with $c_{1:r}$ in place of $c$, and $\widehat V_r$ is the residual spectral energy of prediction
influence at $J'$ validation design points $x'_1,\ldots,x'_{J'}$, drawn
independently of the representative points. Let
$\tilde b_\ell\in\mathbb R^{M_0}$ collect the values of $b_{x'_\ell}$ at
the $M_0$ pilot frequencies, and let $P_r$ be the orthogonal projection in
$\mathbb R^{M_0}$ onto the span of the first $r$ calibration functions
evaluated at the same frequencies. Then
\[
\widehat V_r=\frac1{J'}\sum_{\ell=1}^{J'}\frac{\|(I-P_r)\tilde b_\ell\|^2}{M_0-r},
\]
where, for the hybrids with MM, the Gaussian covariance-score directions
are also projected out and the degrees of freedom are reduced accordingly.
The second factor equals $m/\mathrm{ESS}_r$, where
$\mathrm{ESS}_r=(\sum_jw_{j,r}^{\mathrm{ET}})^2/\sum_j(w_{j,r}^{\mathrm{ET}})^2$
is the effective sample size, and penalises weight concentration. This is a proxy for frequency-approximation error, not
an unbiased estimate of test prediction MSE. The same pilot frequencies
are used to construct the directions and evaluate the proxy, while
prediction test points are excluded from selection. A conservative
sensitivity rule selects the largest feasible rank with
$\mathrm{ESS}_r\ge\tau m$; we use $\tau=0.9$ for landmark checks.
Adaptive-rank and ET performance are evaluated empirically rather than
by the fixed-rank projection theorem.

For full kernel ridge regression we use $120$ representative design points, an independent set of $120$ validation design points, and $1600$ pilot frequencies to estimate the principal directions of prediction influence and its residual energy. Both sets of design points are sampled independently from $N(0,I_5)$. The candidate ranks are $\{4,8,12,16,20,24,28,32,40,48,56,64\}$, and the number of sampled frequencies is $m\in\{50,100,200,400,800\}$ (giving $2m$ real features). We compare ordinary random Fourier features (RFF), Gaussian moment matching (MM), closed-form TCRF, TCRF-ET, and their hybrids with MM over $200$ independent replications. Table~\ref{tab:krr-et} reports RMSE relative to exact KRR prediction; Table~\ref{tab:krr-truth} additionally reports prediction error relative to the true regression function. The two targets answer different questions and must not be conflated.

\begin{table}[t]
\centering
\caption{Kernel ridge regression approximation error over 200 replications. The RFF column gives the mean RMSE with respect to exact KRR; the remaining columns give the ratio of the mean RMSE to RFF, with the mean selected rank in square brackets. TCRF and MM--TCRF use the closed-form weights with rank rule \eqref{eq:rank-rule}; TCRF-ET and MM--TCRF-ET use exponential-tilting weights with the pilot-proxy rank rule \eqref{eq:et-rank-rule}. Lower is better.}
\label{tab:krr-et}
\medskip
\footnotesize
\setlength{\tabcolsep}{4pt}
\begin{tabular}{rcccccc}
\toprule
$m$ & RFF RMSE & MM & TCRF & MM--TCRF & TCRF-ET & MM--TCRF-ET \\
\midrule
\multicolumn{7}{l}{$\lambda = 0.02$} \\
\midrule
50 & 0.1054 & 0.782 & 0.965 [3.5] & 0.739 [7.3] & 0.886 [10.9] & 0.691 [13.6] \\
100 & 0.0768 & 0.767 & 0.894 [6.5] & 0.673 [11.4] & 0.615 [23.8] & 0.529 [25.8] \\
200 & 0.0558 & 0.755 & 0.777 [10.8] & 0.574 [16.7] & 0.440 [44.6] & 0.388 [49.0] \\
400 & 0.0388 & 0.771 & 0.605 [18.4] & 0.479 [23.8] & 0.329 [63.5] & 0.311 [63.8] \\
800 & 0.0273 & 0.775 & 0.476 [27.5] & 0.406 [33.5] & 0.298 [64.0] & 0.287 [64.0] \\
\midrule
\multicolumn{7}{l}{$\lambda$ by five-fold cross-validation (median $2.5\times 10^{-4}$)} \\
\midrule
50 & 0.1794 & 0.901 & 0.995 [1.9] & 0.901 [2.6] & 1.010 [6.4] & 0.911 [6.2] \\
100 & 0.1266 & 0.925 & 0.996 [2.8] & 0.921 [3.8] & 1.008 [14.0] & 0.933 [14.9] \\
200 & 0.0936 & 0.925 & 0.977 [4.5] & 0.901 [5.5] & 0.932 [28.1] & 0.865 [29.4] \\
400 & 0.0677 & 0.937 & 0.958 [6.1] & 0.894 [7.5] & 0.800 [54.9] & 0.755 [57.6] \\
800 & 0.0492 & 0.943 & 0.927 [9.6] & 0.862 [11.1] & 0.680 [63.9] & 0.654 [63.8] \\
\bottomrule
\end{tabular}
\end{table}

\begin{figure}[t]
\centering
\includegraphics[width=.96\linewidth]{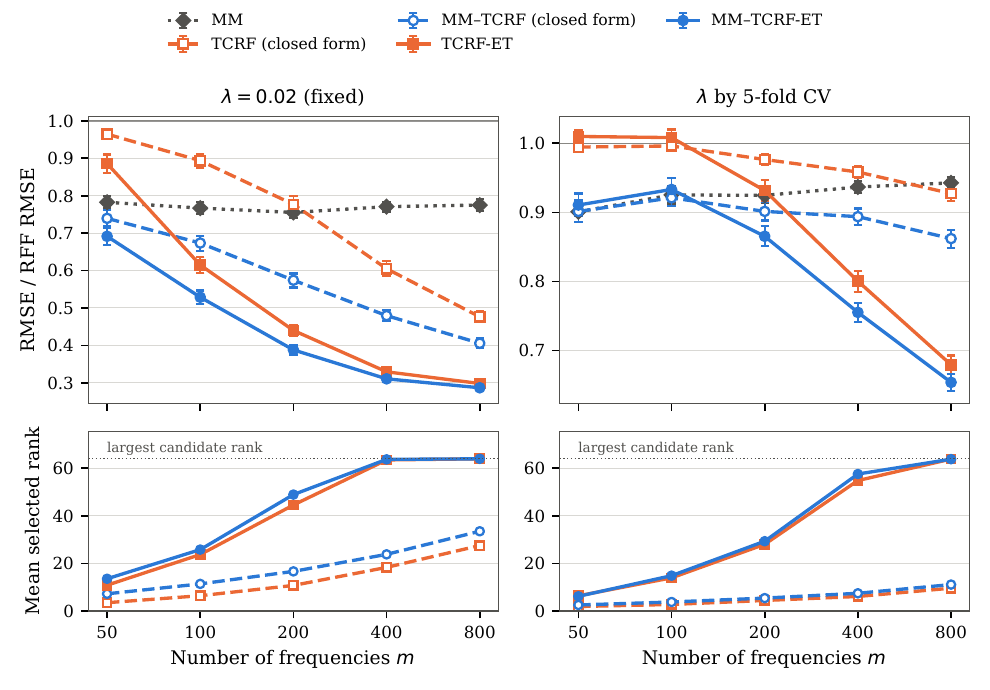}
\caption{KRR approximation RMSE relative to RFF under the fixed and
cross-validated ridge parameters, comparing closed-form and ET
calibration; bars are paired-bootstrap 95\% intervals. The candidate-rank
grid ends at $64$; ET often selects this boundary for the larger
frequency budgets.}
\label{fig:krr-publication}
\end{figure}

Under the fixed $\lambda=0.02$ (Table~\ref{tab:krr-et} and
Figure~\ref{fig:krr-publication}), the ET hybrid reaches an
approximation-RMSE ratio of $0.287$ to RFF at $m=800$, compared
with $0.406$ for the positive closed-form hybrid, while the mean selected
ranks are $64.0$ and $33.5$, respectively. Positive ET weights therefore
allow substantially more task directions to be used without introducing
negative feature coefficients. This comparison is not an isolated
fixed-rank effect of the weight formula: a diagnostic signed
closed-form fit, which does not guarantee a positive-semidefinite
random-feature kernel, reaches a ratio of $0.293$ at mean rank $64$.
The main finite-sample ET benefit here is access to a richer
\emph{positive-weight} calibration space.

With $\lambda$ chosen by cross-validation, the relative gains decrease,
particularly at small $m$. At $m=50$ and $100$, the ET hybrid ratios
$0.911$ and $0.933$ are slightly worse than those of the closed-form
hybrid ($0.901$ and $0.921$). From $m=200$ onward ET improves the
approximation, reaching $0.654$ rather than $0.862$ at $m=800$.
At $m=800$ the mean ET hybrid rank is $63.8$, versus $11.1$ for closed form.
Thus positivity, rank adaptation, and the nonlinear weights jointly
affect the reported improvement; the fixed-rank projection theorem
does not quantify their separate contributions. The stand-alone ET
method improves on MM from $m=100$ under the fixed ridge, but only from
$m=400$ under the cross-validated ridge. The rank ceiling of $64$ is active at large
$m$, motivating the separate extended-grid sensitivity analysis
in Table~\ref{tab:krr-rules} and Figure~\ref{fig:krr-et-extgrid}.

\begin{table}[t]
\centering
\caption{Statistical error: mean RMSE against the true regression function. The block headings give the error of exact KRR; the entries include the additional frequency-approximation error.}
\label{tab:krr-truth}
\medskip
\small
\begin{tabular}{rcccc}
\toprule
$m$ & RFF & MM & MM--TCRF & MM--TCRF-ET \\
\midrule
\multicolumn{5}{l}{$\lambda = 0.02$; exact KRR 0.6791} \\
\midrule
50 & 0.7016 & 0.6895 & 0.6954 & 0.6924 \\
100 & 0.6912 & 0.6855 & 0.6860 & 0.6821 \\
200 & 0.6855 & 0.6820 & 0.6819 & 0.6780 \\
400 & 0.6824 & 0.6802 & 0.6797 & 0.6780 \\
800 & 0.6814 & 0.6799 & 0.6791 & 0.6784 \\
\midrule
\multicolumn{5}{l}{$\lambda$ by five-fold cross-validation; exact KRR 0.2733} \\
\midrule
50 & 0.3136 & 0.3026 & 0.3021 & 0.3015 \\
100 & 0.2922 & 0.2883 & 0.2874 & 0.2835 \\
200 & 0.2840 & 0.2819 & 0.2801 & 0.2732 \\
400 & 0.2788 & 0.2784 & 0.2759 & 0.2729 \\
800 & 0.2762 & 0.2760 & 0.2742 & 0.2727 \\
\bottomrule
\end{tabular}
\end{table}

The fixed ridge leads to mean prediction RMSE $0.6791$ for exact KRR,
whereas the cross-validated ridge reduces this to $0.2733$ (median
$\lambda=2.5\times10^{-4}$; Table~\ref{tab:krr-truth}). Under cross-validation, at $m=200$
the ET hybrid has RMSE $0.2732$, matching exact KRR
within Monte Carlo uncertainty (paired difference $-0.0001$, standard error $0.0008$); no superiority to exact KRR is implied.
Under the fixed ridge the ET hybrid lies slightly below exact KRR for $m\ge200$
(paired differences about $-0.001$); this small effect concerns an
over-regularised reference fit and is not interpreted as an improvement over exact KRR.
For $\lambda=0.02$, frequency approximation changes the final
prediction error relatively little because the regularised reference
itself has substantial error. The importance of approximation for
prediction depends on ridge selection and frequency budget.
Conversely, a large relative reduction in
approximation RMSE need not translate proportionally into test
prediction improvement. For the finite-dimensional calibration subproblem,
solving ET across candidate ranks took about $14$--$72$ milliseconds per fit,
versus $1.5$--$4$ milliseconds for closed form (Appendix~\ref{supp:numerical});
these local timings do not constitute an equal-wall-time comparison.

\subsection{Landmark kernel ridge regression}

We next replace the full exact influence used to construct the calibration space by a small exact KRR problem on $n_L\in\{30,50,100\}$ training landmarks. We use $120$ representative and $120$ distinct validation design points, $1400$ pilot frequencies, $m\in\{100,200,400,800\}$, and candidate ranks $\{2,4,6,8,10,12,16,20,24,32\}$. Each configuration is repeated $150$ times. The full exact KRR serves as the approximation benchmark and, in the cross-validated ridge runs, selects the shared reference $\lambda$; it is not used to construct the landmark influence directions. This design tests whether directions learned from a small KRR problem remain useful for approximating full-sample predictions.

\begin{table}[t]
\centering
\caption{Landmark hybrid relative to moment matching over 150 paired replications: ratio of the mean RMSE of MM--Landmark-TCRF to that of MM, with closed-form weights (CF) and exponential-tilting weights (ET). Lower is better.}
\label{tab:landmark-et}
\medskip
\small
\begin{tabular}{rcccccc}
\toprule
 & \multicolumn{2}{c}{$n_L=30$} & \multicolumn{2}{c}{$n_L=50$} & \multicolumn{2}{c}{$n_L=100$} \\
\cmidrule(lr){2-3}\cmidrule(lr){4-5}\cmidrule(lr){6-7}
$m$ & CF & ET & CF & ET & CF & ET \\
\midrule
\multicolumn{7}{l}{$\lambda = 0.02$} \\
\midrule
100 & 0.964 & 1.011 & 0.943 & 0.905 & 0.923 & 0.807 \\
200 & 0.935 & 0.909 & 0.885 & 0.811 & 0.838 & 0.698 \\
400 & 0.863 & 0.825 & 0.820 & 0.749 & 0.752 & 0.655 \\
800 & 0.834 & 0.802 & 0.776 & 0.733 & 0.679 & 0.637 \\
\midrule
\multicolumn{7}{l}{$\lambda$ by five-fold cross-validation} \\
\midrule
100 & 0.998 & 1.102 & 0.997 & 1.095 & 0.997 & 1.042 \\
200 & 0.993 & 1.111 & 0.989 & 1.106 & 0.992 & 1.053 \\
400 & 0.983 & 1.019 & 0.977 & 1.007 & 0.981 & 0.974 \\
800 & 0.983 & 0.967 & 0.976 & 0.954 & 0.962 & 0.920 \\
\bottomrule
\end{tabular}
\end{table}

\begin{figure}[t]
\centering
\includegraphics[width=.91\linewidth]{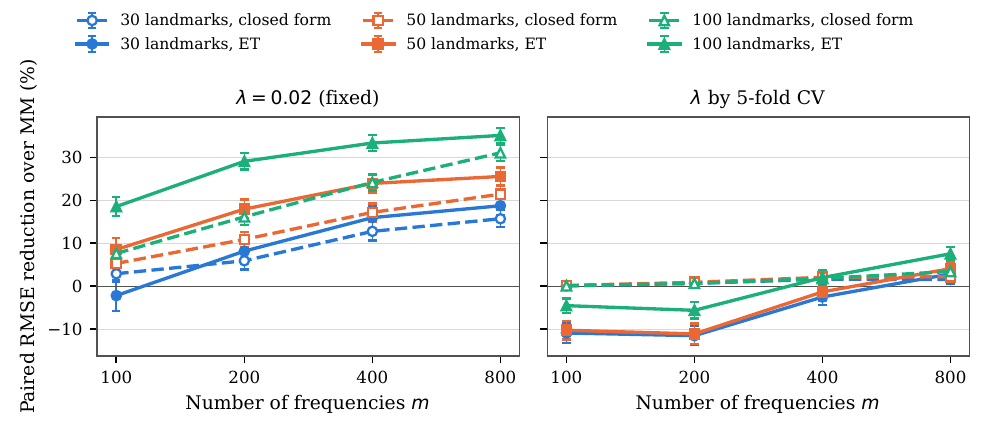}
\caption{Paired percentage reduction in landmark-hybrid approximation
RMSE relative to MM over $150$ replications, with 95\% intervals;
positive values favour the hybrid. The two ridge choices reveal that ET
gains under the fixed $\lambda=0.02$ setting need not carry over to the
cross-validated full-sample prediction task.}
\label{fig:landmark-publication}
\end{figure}

Under $\lambda=0.02$ (Table~\ref{tab:landmark-et} and Figure~\ref{fig:landmark-publication}),
the positive ET hybrid improves on closed form except at the smallest budget
($n_L=30$, $m=100$), and it generally improves with the landmark budget. The improvement over closed form is not monotone in $m$; at $n_L=100$ the
closed-form/MM ratios $0.923$, $0.838$, $0.752$, and $0.679$ become
$0.807$, $0.698$, $0.655$, and $0.637$ with ET for
$m=100,200,400,800$, respectively. These results demonstrate
that a low-dimensional exact subproblem can supply useful directions
without solving the full exact problem to construct the pilot.
The cross-validated ridge reveals an important qualification.
Closed-form landmark calibration has little effect, with ratios
approximately $0.96$--$1.00$ relative to MM. ET with the pilot
proxy \eqref{eq:et-rank-rule} can worsen approximation by about
$4$--$11\%$ at $m\le200$, even though its weights are positive and
empirically balanced. The pilot influence describes the
\emph{landmark} KRR task, and its residual-energy proxy need not
identify the directions that best approximate the full-data
prediction map. In a separate $100$-replication check, imposing the
conservative ESS constraint $\mathrm{ESS}_r\ge0.9m$ largely removes
the small-budget losses: ratios are typically $0.98$--$1.01$
for $m\le200$, while at $n_L=100$, $m=800$ they reach about
$0.92$. Independently cross-validating $\lambda$ within the
landmark subproblem has only modest additional benefit.
Table~\ref{tab:landmark-remedy} gives the detailed sensitivity comparison.
Thus useful landmark
directions are available, but their gain is conditional on
task alignment and sufficiently conservative rank selection.

\subsection{Kernel Stein discrepancy}

\begin{figure}[t]
\centering
\includegraphics[width=.94\linewidth]{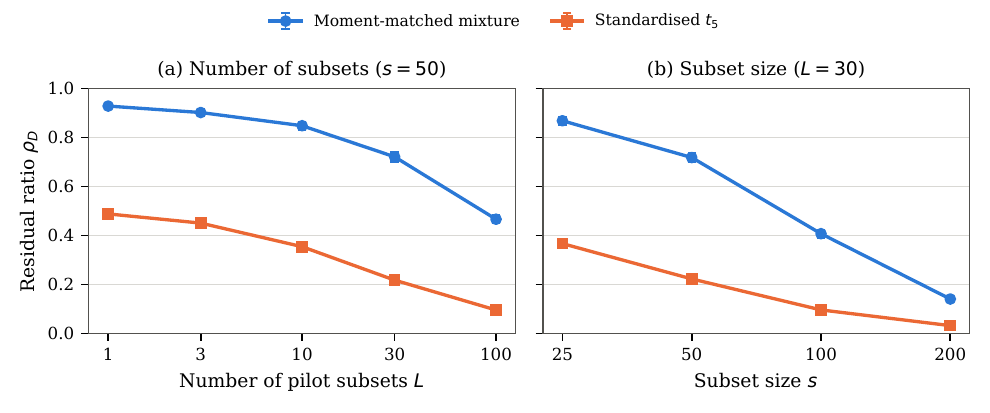}
\caption{Recovery of the KSD zero-variance direction with independent pilot subsets. The residual ratio $\rho_D$ is shown against the number of pilot subsets $L$ at fixed subset size $s=50$ (left) and against subset size $s$ at fixed $L=30$ (right). Error bars are 95\% Monte Carlo intervals over $100$ replications.}
\label{fig:ksd-publication}
\end{figure}

We first examine the subset-recovery mechanism of \eqref{eq:subset-identity} and Theorem~\ref{prop:ksd-blocks}. The target is $N(0,I_d)$ with $d=10$, $n=500$ and Gaussian bandwidth $\sigma=\sqrt d$, and the population frequency projection is approximated on $3000$ frequencies. We use two moment-matched alternatives that are deliberately difficult for low-order moment correction: a symmetric two-component Gaussian mixture with $X_1=0.6S+0.8\varepsilon_1$ and $X_j=\varepsilon_j$ for $j=2,\ldots,d$, where $S$ is Rademacher and $\varepsilon\sim N(0,I_d)$ independently, and a standardised multivariate $t_5$ distribution with identity covariance. Each point in Figure~\ref{fig:ksd-publication} is averaged over $100$ independent replications. The pilots in this diagnostic are independent simple random samples, the design of \eqref{eq:independent-variance}. The purpose is to examine the recovery mechanism rather than the computational cost, and the larger configurations use more pair evaluations than the exact statistic, which is the situation that the disjoint-block design of Theorem~\ref{prop:ksd-blocks} is meant to avoid.

With $s=50$, increasing the number of subsets from $L=1$ to $100$ decreases the residual ratio from $0.929$ to $0.467$ for the moment-matched mixture and from $0.488$ to $0.096$ for the standardised $t_5$. At fixed $L=30$, increasing the subset size from $25$ to $200$ decreases the same ratio from $0.869$ to $0.141$ for the mixture and from $0.367$ to $0.032$ for $t_5$. Both sweeps agree qualitatively with \eqref{eq:independent-variance}. At fixed $s$ the subset-noise variance decreases as $1/L$, while the observed residual ratio decreases slowly at first, consistent with the dependence of the exact residual on noise--signal alignment (Appendix Proposition~\ref{prop:exact-residual}). At fixed $L$, increasing $s$ reduces both the linear and the degenerate terms and is more effective per subset in these experiments, at a cost that grows as $s^2$. The standardised $t_5$ alternative is uniformly easier, which in terms of Theorem~\ref{prop:ksd-blocks} indicates a larger variance of the full-data integrand relative to the pair noise. Full summaries with Monte Carlo standard errors are given in Table~\ref{tab:ksd-detail}.

The residual ratios in the preceding diagnostics measure the recovery of the
frequency integrand, rather than the finite-frequency approximation error itself.
We therefore compare the random-feature estimators directly against the exact
empirical squared V-KSD computed by the analytic Stein kernel over all $n^2$
data pairs. The target is $p=N(0,I_{10})$, with Gaussian kernel bandwidth
$\sigma=\sqrt{10}$. We generate data either from the null $N(0,I_{10})$ or from the
fixed alternative $N(0.5e_1,I_{10})$.
For each scenario, we consider $n\in\{1000,2000,5000\}$, subset sizes
$s\in\{50,100\}$, and $m\in\{50,100,200\}$ sampled frequencies.
All configurations have $100$ independent datasets and eight independent
frequency resamples per dataset; comparisons between methods use the same
frequencies on the same dataset. The exact squared V-KSD is calculated once
per dataset, without frequency approximation. Pilot subsets are sampled once
per dataset and reused for its eight frequency replicates. The random seed
for this experiment is $20261024$.

We compare ordinary random Fourier features (RFF), calibration using only the
centred isotropic scale score $q(\omega)=\|\omega\|^2-10/\sigma^2$, and two-dimensional
calibration on $\operatorname{span}\{q,\bar c\}$, where $\bar c$ averages the
centred subset frequency integrands. The pilot subsets are either $L=n/s$
independently sampled subsets of size $s$ or the $L$ disjoint blocks of a
uniform random partition. The expectation needed to centre each subset
integrand is computed exactly as the corresponding subset squared V-KSD,
with $ns$ pair evaluations in either design (for $L=n/s$). Closed-form
weights use \eqref{eq:rank-rule} with candidate ranks $2,1,0$,
$\kappa_{\mathrm{cond}}=10^{-9}$ and $\varepsilon=0.01$, where rank one
contains only $q$. ET uses the largest
of these ranks with the same conditioning check and a converged
positive dual solution, without the minimum-weight screen.
Unlike KRR, no pilot-proxy rank rule is used in this rank-two
KSD comparison. We measure the squared error relative to the exact empirical
squared V-KSD on each dataset and then compute RMSE across datasets and frequency
replications. Confidence intervals for RMSE ratios use a paired bootstrap
that resamples entire datasets, not individual frequency repetitions.
Because the exact centring constants cost $ns$ pair evaluations, comparable
to the cost of $s$ additional frequencies, Table~\ref{tab:ksd-et} also reports
RFF with $m+s$ frequencies as an equal-budget baseline.

\begin{table}[t]
\centering
\caption{Frequency-approximation RMSE relative to the exact empirical squared V-KSD ($s=100$): ratio to ordinary RFF with $m$ frequencies over 100 datasets and eight frequency draws per dataset. RFF ($m+s$) spends the pilot budget of $ns$ kernel evaluations on $s$ additional frequencies. CF denotes the closed-form weights with rank rule \eqref{eq:rank-rule}; ET denotes exponential-tilting weights. Square brackets give paired-bootstrap 95\% intervals for Disjoint ET over RFF.}
\label{tab:ksd-et}
\medskip
\footnotesize
\setlength{\tabcolsep}{4pt}
\begin{tabular}{rrccccccc}
\toprule
$m$ & $n$ & RFF ($m+s$) & Scale score & Indep.\ CF & Indep.\ ET & Disjoint CF & \multicolumn{2}{c}{Disjoint ET} \\
\midrule
\multicolumn{9}{l}{Null} \\
\midrule
50 & 1000 & 0.574 & 0.996 & 0.959 & 0.961 & 0.962 & 0.957 & [0.922, 0.994] \\
50 & 2000 & 0.578 & 0.977 & 0.961 & 0.949 & 0.978 & 0.971 & [0.935, 1.005] \\
50 & 5000 & 0.608 & 0.989 & 0.999 & 0.994 & 0.974 & 0.967 & [0.939, 0.997] \\
200 & 1000 & 0.810 & 0.981 & 0.924 & 0.923 & 0.938 & 0.938 & [0.913, 0.963] \\
200 & 2000 & 0.806 & 0.976 & 0.951 & 0.950 & 0.941 & 0.939 & [0.916, 0.964] \\
200 & 5000 & 0.821 & 0.982 & 0.974 & 0.974 & 0.967 & 0.967 & [0.942, 0.995] \\
\midrule
\multicolumn{9}{l}{Mean shift} \\
\midrule
50 & 1000 & 0.596 & 0.492 & 0.334 & 0.251 & 0.293 & 0.210 & [0.195, 0.225] \\
50 & 2000 & 0.574 & 0.427 & 0.299 & 0.218 & 0.267 & 0.167 & [0.155, 0.179] \\
50 & 5000 & 0.573 & 0.382 & 0.228 & 0.142 & 0.182 & 0.105 & [0.098, 0.112] \\
200 & 1000 & 0.824 & 0.363 & 0.248 & 0.245 & 0.195 & 0.193 & [0.180, 0.207] \\
200 & 2000 & 0.824 & 0.351 & 0.211 & 0.207 & 0.169 & 0.162 & [0.152, 0.172] \\
200 & 5000 & 0.815 & 0.296 & 0.143 & 0.138 & 0.115 & 0.103 & [0.096, 0.110] \\
\bottomrule
\end{tabular}
\end{table}

\begin{figure}[t]
\centering
\includegraphics[width=.96\linewidth]{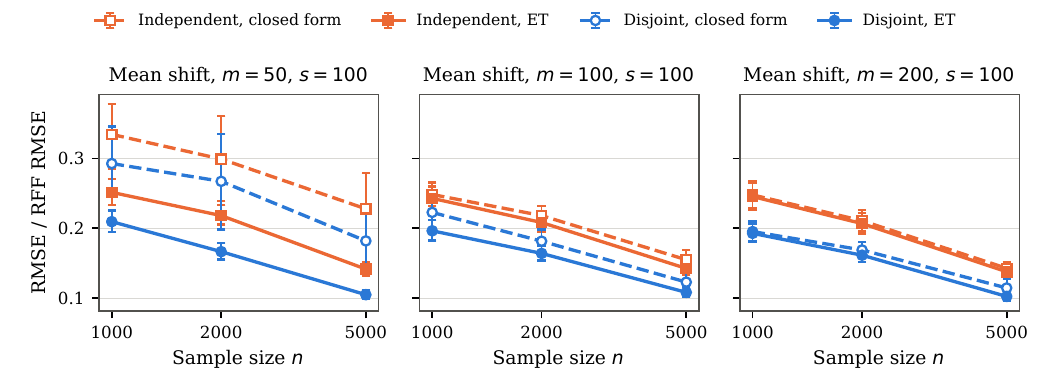}
\caption{Exact empirical squared V-KSD approximation error under the
fixed mean shift: closed-form and ET calibration with independent and
disjoint-block pilots ($s=100$). Error ratios are relative to uncalibrated
RFF on paired datasets and frequencies; bars are paired-bootstrap 95\%
intervals. Null results are reported in Table~\ref{tab:ksd-et}.}
\label{fig:ksd-exact-validation}
\end{figure}

Under the fixed mean shift (Table~\ref{tab:ksd-et} and
Figure~\ref{fig:ksd-exact-validation}), the disjoint-block ET method improves
most at the smaller frequency budget. With $s=100$, $m=50$,
the disjoint/RFF ratios at $n=1000,2000,5000$ are
$0.293$, $0.267$, and $0.182$ for closed form, compared with
$0.210$, $0.167$, and $0.105$ for ET.
Some closed-form samples fall back below rank two because of its
positivity screen, whereas the ET solver maintains rank two in
the reported small-budget cases. This is consistent with the
importance of feasible positive-weight calibration, although
the observed loss cannot be attributed solely to rank fallback
without a separate fixed-rank ablation.
At $m=200$, the ET and closed-form disjoint/RFF ratios are
$0.193$ versus $0.195$ at $n=1000$,
$0.162$ versus $0.169$ at $n=2000$, and
$0.103$ versus $0.115$ at $n=5000$.
Thus ET gives an implementation improvement without changing
the fixed-data subset identity or the disjoint-block theorem.
The gain persists at equal budget: RFF with $m+s$ frequencies attains
ratios $0.57$--$0.82$, three to eight times larger than those of the
disjoint ET estimator.

Under the null, ET and closed-form disjoint pilots are nearly identical:
their \emph{mutual} RMSE ratios are close to one, while their ratios
to uncalibrated RFF are about $0.94$--$0.98$. At equal budget, however,
spending the pilot cost on $s$ additional frequencies is more effective:
RFF with $m+s$ frequencies attains ratios $0.57$--$0.82$. The
conversion of pair evaluations into frequencies counts operations rather
than wall-clock time. In the timing check of Appendix~\ref{supp:numerical},
at $n=5000$, $s=100$ and $m=200$ the pilot costs about as much as $230$
additional frequencies, so the comparison is unchanged in wall-clock terms,
while the calibrated approximation is about seven times faster than the
exact statistic. This experiment concerns approximation of the empirical statistic,
not the size or power of a resulting goodness-of-fit test.

\section{Discussion}
\label{sec:discussion}

Kernel ridge regression and the kernel Stein discrepancy differ in an instructive way. Because the prediction map involves the inverse of the kernel matrix, a second-order error remains after the first-order term has been cancelled; under complete containment a regime of order $m^{-1}$ appears in which the limit is a signed quadratic form rather than a normal distribution, and the bias and the standard deviation then have the same order. For a fixed near-zero space the rate remains $m^{-1/2}$, and only the leading variance constant is reduced. For the Stein discrepancy the frequency integrand is itself the target fluctuation, so exact containment removes the frequency error completely, and the subset identity turns cheap computations on subsets of the data into a surrogate for the full-data zero-variance direction. The disjoint-block analysis applies the same balancing idea to the data: a random partition balances the linear component of the subset noise exactly, so that the pilot budget is governed by the ratio of the degenerate variance to the variance of the full-data integrand rather than by the sample size.

This analysis also delimits the scope of the Stein discrepancy construction. Under a fixed alternative, the pilot cost relative to the exact statistic vanishes as the sample size grows, the calibrated random-feature discrepancy costs order $(m+s)nd$, and its accuracy gain persists against random features of equal budget. Under the null hypothesis, degenerate-U-statistic scaling makes the subset-noise bound uninformative when $s\ll n$, and the same budget is better spent on additional frequencies. Whether a pilot of linear cost can reduce the frequency error near the null remains open, as does a full evaluation of the size and power of the resulting goodness-of-fit test, in which calibration and pilot construction must be accounted for when generating the null reference distribution.

The regression experiments separate calibration effects from ridge selection. Under a common ridge parameter the positive ET weights allow larger calibration ranks than the closed-form positivity screen, reducing the approximation-RMSE ratio of the hybrid at $m=800$ from $0.406$ to $0.287$ for $\lambda=0.02$ and from $0.862$ to $0.654$ under cross-validation; since the adaptive rank rules differ, these comparisons measure complete practical procedures rather than a fixed-rank effect of the weight family. Cross-validation reduces the true-function prediction RMSE of exact KRR from $0.6791$ to $0.2733$, so the prediction value of a frequency-approximation gain depends on regularisation and budget. The landmark version shows that an aggressively selected ET rank can harm full-sample prediction at small $m$, while a conservative ESS screen limits this loss; calibration gains thus depend on alignment between the pilot influence and the final prediction task. The large-sample check in Appendix~\ref{sec:supp-large-krr} charges the pilot to the frequency budget but is limited to closed-form weights and a fixed ridge parameter.

The theory proved here covers fixed-rank, exactly centred calibration with closed-form weights. It requires square-integrable calibration functions, which is automatic for kernel ridge regression, whose controls are bounded, but for the Stein discrepancy requires finite fourth moments of the spectral law; this excludes, for example, Mat\'ern kernels of smoothness at most two. A rigorous treatment of adaptive or growing rank, of approximately centred Nystr\"om--Taylor controls, and of the second-order regime of the ET weights is left for future work, as are a comprehensive equal-wall-time accuracy frontier and further models.

\bibliographystyle{plainnat}
\bibliography{cancellation_principle_random_features}

\clearpage
\appendix
\section*{Appendix}
\numberwithin{equation}{section}
\numberwithin{figure}{section}
\numberwithin{table}{section}
\numberwithin{proposition}{section}
\numberwithin{lemma}{section}
\numberwithin{theorem}{section}
\numberwithin{corollary}{section}

\section{Disjoint-block pilots for the kernel Stein discrepancy}
\label{supp:ksd}

Throughout this section the dataset $D$ is fixed, $n\ge4$, $2\le s\le n$ and $N=Ls\le n$, and all statements hold pointwise in the frequency $\omega$ before integration against $q_0$. We use the notation of the Stein discrepancy part: $\kappa_{ij}(\omega)=\psi_\omega(X_i)^\top\psi_\omega(X_j)-h_p(X_i,X_j)$ for $i\neq j$, $a=a_{n,s}$, $b=b_{n,s}$, and $\zeta_{ij}$ for the number of pilot subsets containing both $i$ and $j$ (so that $\zeta_{ij}\in\{0,1\}$ in the disjoint-block design). Since $h_p(x,x)=\|s_p(x)\|^2+\mathbb{E}_\omega\|\omega\|^2$ and $\|\psi_\omega(x)\|^2=\|s_p(x)\|^2+\|\omega\|^2$, the diagonal terms of $G_S$ centre to $q$, and therefore
\[
c_S=\frac1sq+\frac1{s^2}\sum_{i\neq j\in S}\kappa_{ij},
\quad
u_D=\frac1nq+\frac1{n^2}\sum_{i\neq j}\kappa_{ij},
\quad
\bar c=\frac1sq+\frac1{Ns}\sum_{i\neq j}\zeta_{ij}\kappa_{ij},
\]
where the last identity uses $Ls^2=Ns$.

\begin{lemma}[Finite-population Hoeffding decomposition]
\label{lem:hoeffding}
Let $(\kappa_{ij})_{i\ne j}$ be a symmetric real array with $n\ge3$, and put
\[
\bar\kappa=\frac1{n(n-1)}\sum_{i\ne j}\kappa_{ij},
\quad
\bar\kappa_i=\frac1{n-1}\sum_{j\ne i}\kappa_{ij}.
\]
Then
\[
\mu=\bar\kappa,\quad
\alpha_i=\frac{n-1}{n-2}(\bar\kappa_i-\bar\kappa),\quad
\gamma_{ij}=\kappa_{ij}-\mu-\alpha_i-\alpha_j
\]
give the unique decomposition
\[
\kappa_{ij}=\mu+\alpha_i+\alpha_j+\gamma_{ij},\quad
\sum_i\alpha_i=0,\quad
\gamma_{ij}=\gamma_{ji},\quad
\sum_{j\ne i}\gamma_{ij}=0.
\]
The decomposition also satisfies, for arbitrary real $e,a_1,\ldots,a_n$,
\[
\sum_{i\ne j}\gamma_{ij}(e+a_i+a_j)=0,
\quad
\sum_{i\ne j}\kappa_{ij}^2
=n(n-1)\mu^2+2(n-2)\sum_i\alpha_i^2
+\sum_{i\ne j}\gamma_{ij}^2.
\]
\end{lemma}

\begin{proof}
Since $\sum_i\bar\kappa_i=n\bar\kappa$, the $\alpha_i$ sum to zero. For fixed $i$, using $\sum_{j\neq i}\alpha_j=-\alpha_i$,
\[
\sum_{j\neq i}\gamma_{ij}=(n-1)(\bar\kappa_i-\mu)-(n-1)\alpha_i+\alpha_i=(n-1)(\bar\kappa_i-\mu)-(n-2)\alpha_i=0 .
\]
For uniqueness, suppose $e+a_i+a_j+g_{ij}=0$ for all $i\neq j$ with $\sum_ia_i=0$ and $\sum_{j\neq i}g_{ij}=0$. Summing over $j\neq i$ gives $(n-1)e+(n-2)a_i=0$ for every $i$, and summing over $i$ gives $e=0$, hence $a_i=0$ because $n\ge3$, and then $g_{ij}=0$. The orthogonality relation follows from $\sum_{i\neq j}\gamma_{ij}a_i=\sum_ia_i\sum_{j\neq i}\gamma_{ij}=0$, its symmetric counterpart, and $\sum_{i\neq j}\gamma_{ij}=0$. For the last identity, the cross term between $\mu$ and the linear part is $2(n-1)\mu\sum_i\alpha_i=0$, and $\sum_{i\neq j}(\alpha_i+\alpha_j)^2=2(n-1)\sum_i\alpha_i^2+2\sum_{i\neq j}\alpha_i\alpha_j=2(n-2)\sum_i\alpha_i^2$.
\end{proof}

Applied to the pair functions $\kappa_{ij}(\omega)$ for each $\omega$, the lemma defines $\mu(\omega)$, $\alpha_i(\omega)$ and $\gamma_{ij}(\omega)$ as fixed linear combinations of the $\kappa_{ij}(\omega)$. In particular, $u_D=n^{-1}q+((n-1)/n)\mu$, and the quantities $\bar A$ and $\bar\Gamma$ are finite whenever the Stein features have finite fourth moments under the spectral law, which holds for the Gaussian law because the pair functions grow at most quadratically in the frequency. The decomposition is a finite-population analogue of the Hoeffding decomposition of a U-statistic: the linear terms play the role of the H\'ajek projection, and the doubly centred terms that of the degenerate kernel, which is what makes the two contributions in Theorem~\ref{prop:ksd-blocks} separate cleanly.

\begin{lemma}[Joint selection probabilities]
\label{lem:selection}
In the disjoint-block design, let $T$ be the union of the blocks, and for
$i\ne j$ let $\zeta_{ij}=\mathbf 1\{i,j\text{ are in the same block}\}$.
Then, for distinct indices $i,j,k,l$,
\[
p_2=P(\zeta_{ij}=1)=\frac{N(s-1)}{n(n-1)},
\quad
p_3=P(\zeta_{ij}=\zeta_{ik}=1)
=\frac{N(s-1)(s-2)}{n(n-1)(n-2)},
\]
and
\[
p_4=P(\zeta_{ij}=\zeta_{kl}=1)
=\frac{N(s-1)\left((s-2)(s-3)+(N-s)(s-1)\right)}{n(n-1)(n-2)(n-3)}.
\]
The union $T$ is a simple random sample of size $N$, with
$\sum_{j\ne i}\zeta_{ij}=(s-1)\mathbf1\{i\in T\}$, and
$P(k\in T,\zeta_{ij}=1)$ depends on the indices only
through whether $k\in\{i,j\}$.
\end{lemma}

\begin{proof}
The blocks are consecutive segments of the first $N$ entries of a uniformly random permutation, so $T$ is a simple random sample of size $N$, every unit of $T$ has exactly $s-1$ blockmates, and units outside $T$ have none. Given $i\in T$, which has probability $N/n$, the other $s-1$ members of its block form a simple random sample from the remaining $n-1$ units, which gives the first two probabilities. Given that $i$ and $j$ share a block $B_0$, the other $n-2$ units occupy the remaining $n-2$ positions of the permutation in uniformly random order: $s-2$ positions in $B_0$, $(L-1)s=N-s$ positions in the other blocks, and $n-N$ positions outside $T$. The ordered pair $(k,l)$ therefore occupies a uniformly distributed ordered pair of distinct positions among these $n-2$, of which there are $(n-2)(n-3)$. Of these pairs, $(s-2)(s-3)$ lie in $B_0$ and $(L-1)s(s-1)=(N-s)(s-1)$ lie within one of the other blocks, so that $k$ and $l$ share a block with conditional probability
\[
\frac{(s-2)(s-3)+(N-s)(s-1)}{(n-2)(n-3)} .
\]
Multiplying by $p_2$ gives $p_4$; the count is valid for every $2\le s\le n$. The last statement follows from exchangeability of the units under a uniformly random permutation.
\end{proof}

\begin{proof}[Proof of Theorem~\ref{prop:ksd-blocks}]
In both designs $\mathbb{E}\zeta_{ij}=p_2$, with $p_2$ as in Lemma~\ref{lem:selection}; in the independent design this is $L$ times the probability $s(s-1)/(n(n-1))$ that a single subset contains both units. Hence $\mathbb{E}\bar c=s^{-1}q+((s-1)/s)\mu$, and since $a/n+b=1/s$ and $a(n-1)/n=(s-1)/s$, this equals $a u_D+b q$.

Consider the disjoint-block design and put $\varepsilon=\bar c-\mathbb{E}\bar c=(Ns)^{-1}\sum_{i\neq j}(\zeta_{ij}-p_2)\kappa_{ij}$. Insert the decomposition of Lemma~\ref{lem:hoeffding}. By Lemma~\ref{lem:selection}, $\sum_{i\neq j}(\zeta_{ij}-p_2)=N(s-1)-n(n-1)p_2=0$, so $\mu$ contributes nothing. The linear part contributes
\[
\sum_{i\neq j}(\zeta_{ij}-p_2)(\alpha_i+\alpha_j)=2\sum_i\alpha_i\sum_{j\neq i}(\zeta_{ij}-p_2)=2(s-1)\sum_{i\in T}\alpha_i ,
\]
because $\sum_i\alpha_i=0$, and the degenerate part contributes $\sum_{i\neq j}\zeta_{ij}\gamma_{ij}$, because $\sum_{i\neq j}\gamma_{ij}=0$. Thus $Ns\,\varepsilon=\Lambda_1+\Lambda_2$ with $\Lambda_1=2(s-1)\sum_{i\in T}\alpha_i$ and $\Lambda_2=\sum_{i\neq j}\zeta_{ij}\gamma_{ij}$. Since $T$ is a simple random sample of size $N$ and the $\alpha_i$ sum to zero,
\[
\mathbb{E}\Lambda_1^2=4(s-1)^2\frac{N(n-N)}{n(n-1)}\sum_i\alpha_i^2 .
\]
The cross moment is $\mathbb{E}(\Lambda_1\Lambda_2)=2(s-1)\sum_k\sum_{i\neq j}\alpha_k\gamma_{ij}P(k\in T,\zeta_{ij}=1)$. By the last part of Lemma~\ref{lem:selection} the probability takes one value when $k\in\{i,j\}$ and another when $k\notin\{i,j\}$, and for each $k$ both $\sum_{i\neq j,\,k\in\{i,j\}}\gamma_{ij}=2\sum_{j\neq k}\gamma_{kj}$ and $\sum_{i\neq j,\,k\notin\{i,j\}}\gamma_{ij}$ vanish by double centring, so the cross moment is zero. For $\Lambda_2$,
\[
\mathbb{E}\Lambda_2^2=\sum_{i\neq j}\sum_{k\neq l}P(\zeta_{ij}=\zeta_{kl}=1)\gamma_{ij}\gamma_{kl}.
\]
Fix an ordered pair $(i,j)$ and group the ordered pairs $(k,l)$ by their overlap with $\{i,j\}$. The two pairs with $\{k,l\}=\{i,j\}$ contribute $2\gamma_{ij}$ with probability $p_2$; the pairs sharing exactly one index contribute $2\sum_{l\notin\{i,j\}}\gamma_{il}+2\sum_{l\notin\{i,j\}}\gamma_{jl}=-4\gamma_{ij}$ with the common probability $p_3$; and since $\sum_{k\neq l}\gamma_{kl}=0$, the disjoint pairs contribute $2\gamma_{ij}$ with the common probability $p_4$. Therefore $\mathbb{E}\Lambda_2^2=(2p_2-4p_3+2p_4)\sum_{i\neq j}\gamma_{ij}^2$, and the identity $(n-2)(n-3)-2(s-2)(n-3)+(s-2)(s-3)=(n-s)(n-s-1)$ gives
\[
2p_2-4p_3+2p_4=\frac{2N(s-1)\left((n-s)(n-s-1)+(N-s)(s-1)\right)}{n(n-1)(n-2)(n-3)} .
\]
Dividing $\mathbb{E}\Lambda_1^2+\mathbb{E}\Lambda_2^2$ by $(Ns)^2$, integrating against $q_0$, and using $\sum_i\|\alpha_i\|^2=n\bar A$ and $\sum_{i\neq j}\|\gamma_{ij}\|^2=n(n-1)\bar\Gamma$ gives \eqref{eq:block-variance}.

In the independent design, $\bar c-\mathbb{E}\bar c$ is the average of $L$ independent copies of $c_S-\mathbb{E}c_S$ for a single simple random sample of size $s$, which is the disjoint-block design with $L=1$ and $N=s$. Its variance is therefore given by \eqref{eq:block-variance} with $N=s$, and dividing by $L$ and using $Ls=N$ gives \eqref{eq:independent-variance}. Finally, $a^{-1}(\bar c-bq)\in\mathcal C$ and $u_D-a^{-1}(\bar c-bq)=-a^{-1}(\bar c-\mathbb{E}\bar c)$, so $\operatorname{dist}^2(u_D,\mathcal C)\le a^{-2}\|\bar c-\mathbb{E}\bar c\|^2_{L_2(q_0)}$, and taking expectations completes the proof.
\end{proof}

The exact residual of the subset calibration on $\mathcal C=\operatorname{span}\{\bar c,q\}$ has a simple closed form. It also isolates the component along $q$, which the two-dimensional calibration space reproduces without error, so that only the component of $u_D$ orthogonal to $q$ has to be recovered from the subsets. Let $P^\perp$ denote the orthogonal projection in $L_2(q_0)$ onto the orthogonal complement of $q$, and put $\tilde u=P^\perp u_D$ and $\tilde\varepsilon=P^\perp(\bar c-\mathbb{E}\bar c)$; norms and inner products below are those of $L_2(q_0)$.

\begin{proposition}[Exact residual of the subset calibration]
\label{prop:exact-residual}
If $q\ne0$ and $a\tilde u+\tilde\varepsilon\ne0$, then
\[
\operatorname{dist}^2(u_D,\operatorname{span}\{\bar c,q\})
=
\frac{\|\tilde u\|^2\|\tilde\varepsilon\|^2
-\langle\tilde u,\tilde\varepsilon\rangle^2}
{\|a\tilde u+\tilde\varepsilon\|^2}.
\]
\end{proposition}

\begin{proof}
Since $\mathbb{E}\bar c=au_D+bq$ by \eqref{eq:subset-identity} and $P^\perp q=0$, one has $P^\perp\bar c=a\tilde u+\tilde\varepsilon=:v$, and $\operatorname{span}\{\bar c,q\}$ is the orthogonal sum of $\operatorname{span}\{q\}$ and $\operatorname{span}\{v\}$. The component of $u_D$ along $q$ is therefore reproduced exactly, and the squared distance equals $\|\tilde u\|^2-\langle\tilde u,v\rangle^2/\|v\|^2$. Expanding $\|\tilde u\|^2\|v\|^2-\langle\tilde u,v\rangle^2$ gives $\|\tilde u\|^2\|\tilde\varepsilon\|^2-\langle\tilde u,\tilde\varepsilon\rangle^2$.
\end{proof}

When the realised cross term $\langle\tilde u,\tilde\varepsilon\rangle$
vanishes and $\|\tilde u\|>0$, the residual relative to
$\|\tilde u\|^2$ equals
\[
\frac{\|\tilde\varepsilon\|^2}{a^2\|\tilde u\|^2+\|\tilde\varepsilon\|^2}.
\]
Although $\mathbb{E}_S\langle\tilde u,\tilde\varepsilon\rangle=0$
conditionally on the dataset, the realised cross term need not vanish,
and it cannot be removed from the nonlinear residual formula before
taking expectations; the display above therefore describes the
orthogonal-noise case rather than the expected residual.

For every subset realisation, a consequence that does not require this
orthogonality condition is
\[
\operatorname{dist}^2(u_D,\mathcal C)
\le\min\left\{
\|\tilde u\|^2,\,
\frac{\|\tilde\varepsilon\|^2}{a^2}
\right\}.
\]
The first bound follows because $\mathcal C$ contains $q$; the second
follows by approximating $\tilde u$ with
$a^{-1}P^\perp\bar c=\tilde u+a^{-1}\tilde\varepsilon$.
Consequently, whenever $\|u_D\|>0$,
\[
\mathbb{E}_S\!\left[\rho_D(\mathcal C)\right]
\le\min\left\{
\frac{\|\tilde u\|^2}{\|u_D\|^2},\,
\frac{\mathbb{E}_S\|\tilde\varepsilon\|^2}
{a^2\|u_D\|^2}
\right\},
\]
where the expectation is over the subset design with the dataset fixed.

Because $P^\perp$ and the finite-population Hoeffding decomposition are
linear, Theorem~\ref{prop:ksd-blocks} applies to $\tilde\varepsilon$
after replacing each pair function by its projection onto the orthogonal
complement of $q$. A full partition therefore removes the linear
component of the projected subset noise exactly. Under a fixed alternative
with a spectral signal bounded away from zero and bounded projected pair
moments, the resulting bound decreases at order $1/(ns)$ for $s\ll n$.
Near the null, the smaller spectral signal can make this relative-error
bound uninformative, although this neither establishes a lower bound on
the pilot cost nor rules out variance reduction beyond the isotropic
direction.

\section{First-order expansion for Gaussian random features and moment matching}
\label{supp:mm}

We record the relations used in the framework section, with $\bar Z$, $S_m$ and $\widetilde Z_j=S_m^{-1/2}(Z_j-\bar Z)$ as defined there. For fixed $d$ and fixed $\delta$, as the number of frequencies grows,
\[
\hat k_{\mathrm{MM}}(\delta)
=
\hat k_{\mathrm{RFF}}(\delta)
+\frac{k(\delta)}2\delta^{\top}(S_m-I_d)\delta
+o_p(m^{-1/2}).
\]
The first-order term of the mean correction vanishes because $\mathbb{E}(Z\cos(Z^{\top}\delta))=0$ by symmetry. The estimated-sampler estimator obtained by estimating the mean and covariance parameters of the Gaussian spectral sampler from the same frequency sample has the same first-order term, so that $\hat k_{\mathrm{ES}}(\delta)-\hat k_{\mathrm{MM}}(\delta)=o_p(m^{-1/2})$. Writing $u_\delta(Z)=\cos(Z^{\top}\delta)-k(\delta)$, the residual after removing the Gaussian covariance-score projection is
\[
r_\delta(Z)
=u_\delta(Z)
+\frac{k(\delta)}2\left((Z^{\top}\delta)^2-\|\delta\|^2\right),
\]
and for two lags $\delta$ and $\delta'$ the covariances before and after the projection are
\[
\Gamma_0(\delta,\delta')
=k(\delta)k(\delta')\left(\cosh(\delta^{\top}\delta')-1\right),
\]
\[
\Gamma_{\mathrm{res}}(\delta,\delta')
=k(\delta)k(\delta')
\left(\cosh(\delta^{\top}\delta')-1-\frac12(\delta^{\top}\delta')^2\right).
\]
In particular, as $\|\delta\|\to0$, one has $\Gamma_0(\delta,\delta)=\|\delta\|^4/2+O(\|\delta\|^6)$ and $\Gamma_{\mathrm{res}}(\delta,\delta)=\|\delta\|^8/24+O(\|\delta\|^{10})$. This higher-order cancellation explains the strength of Gaussian moment matching for nearby inputs, but the proposed method does not place its novelty here; the task-derived calibration space is designed for the prediction influence, which also involves lags that are not small.

\paragraph{Overlap of task and Gaussian-score directions.}
Absorb the bandwidth into the lag $\delta$ and write
$g_\delta(Z)=\cos(Z^\top\delta)-k(\delta)$. Because the KRR
influence $b_x$ is a finite linear combination of centred kernel
entries, it may be expressed using a discrete signed measure $\mu_x$
on the lag space as $b_x(Z)=\int g_\delta(Z)\,d\mu_x(\delta)$.
Let $\mathcal Q=\{Z^\top AZ-\operatorname{tr}(A):A=A^\top\}$ be the
Gaussian covariance-score space. The Hermite
projection of the influence onto $\mathcal Q$ is
\[
\Pi_{\mathcal Q}b_x(Z)
=-\frac12\left(Z^\top M_xZ-\operatorname{tr}(M_x)\right),
\quad
M_x=\int k(\delta)\,\delta\delta^\top\,d\mu_x(\delta).
\]
Hence task and covariance-score directions need not be orthogonal;
the joint rank depends on the prediction problem.

\section{Derivation of the task influence for kernel ridge regression}
\label{supp:krr}

We use the standardised frequency $Z\sim N(0,I_d)$ and the Gaussian bandwidth $\sigma$. In particular, $K_{ii'}=k(X_i,X_{i'})$, $(k_x)_i=k(x,X_i)$, and the single-frequency contributions have entries
\[
(A_Z)_{ii'}=\cos\left(\frac{Z^{\top}(X_i-X_{i'})}{\sigma}\right),\quad
(a_Z(x))_i=\cos\left(\frac{Z^{\top}(x-X_i)}{\sigma}\right).
\]
Let $\widehat K=K+E$ and $\hat k_x=k_x+e_x$. Since $R=(K+n\lambda I_n)^{-1}$, the resolvent expansion is
\[
(K+E+n\lambda I_n)^{-1}
=R-RER+O(\|E\|^2),
\]
and therefore
\[
\hat f(x)-f_K(x)
=e_x^{\top}\alpha-k_x^{\top}RE\alpha+O_p(m^{-1}),
\quad \alpha=Ry .
\]
With these definitions, $\mathbb{E}_ZA_Z=K$ and $\mathbb{E}_Za_Z(x)=k_x$. The first-order term is then
\[
\frac1m\sum_{j=1}^m
\left(
(a_{Z_j}(x)-k_x)^{\top}\alpha
-k_x^{\top}R(A_{Z_j}-K)\alpha
\right),
\]
and the influence function is defined as the summand, from which $\mathbb{E}(b_x(Z))=0$ follows. This representation does not require the dense matrix $A_Z$ to be formed. With the paired cosine and sine representation the contribution of one frequency is of rank two, so that the matrix--vector product can be evaluated through two inner products. The dominant cost of the pilot influence matrix is therefore of order $M_0nJ$, where $M_0$ is the number of pilot frequencies and $J$ the number of representative points, and of order $M_0n_LJ$ in the landmark version.

\section{Additional numerical results}
\label{supp:numerical}

This section reports ET and cross-validated-ridge experiments alongside
an independent closed-form-only benchmark. The KRR experiment has $200$
datasets, the landmark experiment has $150$ per configuration, and the
exact empirical KSD benchmark has $100$ datasets with eight frequency
resamples on each dataset. The separate closed-form benchmark uses
independent random seeds and therefore does not support paired
comparisons with the ET/CV tables.

\subsection{Positive ET weights, rank sensitivity and cross-validation}
\label{supp:et-cv}

The KRR experiments use $n=300$, $d=5$, the specified
Gaussian-kernel bandwidth and regression model, and $m=50,100,200,400,800$.
The cross-validated ridge is selected separately in each training
replication by five folds of \emph{exact KRR}, on a 26-point logarithmic
grid from $10^{-6}$ to $10^{-1}$. The median selected value is
$2.5\times10^{-4}$. This is a reference-calibration comparison;
it does not eliminate the need to tune an RF or Nystr\"om method
at acceptable cost in large datasets. All approximation methods
use the same $\lambda$ within a replication.

For each KRR dataset we draw $120$ representative design points
and a disjoint set of $120$ validation design points independently
from $N(0,I_5)$, together with $1600$ pilot frequencies.
The validation points are distinct from the held-out test points;
the pilot frequency sample is shared between fitting the principal
directions and computing the rank proxy $\widehat V_r$.
The ET dual is solved with damped Newton iterations, after
checking candidate feasibility and conditioning. The reported
fit-level timings for the candidate-rank ET solves are
$14$--$72$ ms, compared with $1.5$--$4$ ms for closed form,
on the measured local configuration; these do not include
exact-KRR cross-validation or all pilot construction.
The KRR, landmark and KSD seeds are
$20261022$, $20261023$, and $20261024$, respectively.
The extended rank-grid check uses seed $20261025$.
The arXiv ancillary archive \texttt{TCRF\_reproduction.zip} contains
reproduction scripts, raw CSVs, dataset-level aggregates and plotting
commands for the numerical results.

\begin{table}[t]
\centering
\caption{Sensitivity of MM--TCRF to the weight family and the rank rule: ratio of the mean RMSE to RFF, with the mean selected rank in square brackets. CF is the closed-form rule \eqref{eq:rank-rule}; ET-ESS$_{\tau}$ selects the largest rank whose exponential-tilting weights have effective sample size at least $\tau m$; ET-pMSE minimises the pilot proxy \eqref{eq:et-rank-rule}. $^{\dagger}$The candidate grid is extended by 80, 96 and 112, on an independent set of 100 replications. Lower is better.}
\label{tab:krr-rules}
\medskip
\footnotesize
\setlength{\tabcolsep}{4pt}
\begin{tabular}{rccccc}
\toprule
$m$ & CF & ET-ESS$_{0.5}$ & ET-ESS$_{0.9}$ & ET-pMSE & ET-pMSE$^{\dagger}$ \\
\midrule
\multicolumn{6}{l}{$\lambda = 0.02$} \\
\midrule
50 & 0.739 [7.3] & 0.694 [15.6] & 0.742 [5.7] & 0.691 [13.6] & 0.674 [13.6] \\
100 & 0.673 [11.4] & 0.520 [28.2] & 0.670 [11.0] & 0.529 [25.8] & 0.507 [25.8] \\
200 & 0.574 [16.7] & 0.388 [48.2] & 0.532 [20.1] & 0.388 [49.0] & 0.374 [49.6] \\
400 & 0.479 [23.8] & 0.310 [63.9] & 0.410 [34.0] & 0.311 [63.8] & 0.276 [85.4] \\
800 & 0.406 [33.5] & 0.287 [64.0] & 0.303 [58.0] & 0.287 [64.0] & 0.196 [111.8] \\
\midrule
\multicolumn{6}{l}{$\lambda$ by five-fold cross-validation} \\
\midrule
50 & 0.901 [2.6] & 0.992 [11.3] & 0.895 [2.9] & 0.911 [6.2] & 0.923 [6.2] \\
100 & 0.921 [3.8] & 0.985 [20.4] & 0.915 [6.4] & 0.933 [14.9] & 0.914 [14.4] \\
200 & 0.901 [5.5] & 0.890 [35.3] & 0.868 [13.0] & 0.865 [29.4] & 0.885 [28.7] \\
400 & 0.894 [7.5] & 0.757 [58.8] & 0.810 [23.8] & 0.755 [57.6] & 0.764 [60.2] \\
800 & 0.862 [11.1] & 0.654 [63.8] & 0.712 [42.1] & 0.654 [63.8] & 0.558 [104.3] \\
\bottomrule
\end{tabular}
\end{table}

The KRR rule sensitivity in Table~\ref{tab:krr-rules} separates admissibility from the
nonlinear weight family. The signed closed-form diagnostic
allows higher ranks but may destroy positive semidefiniteness.
For example, at $m=800$ the ratios to RFF for the
positive closed-form, signed closed-form and ET hybrid are,
respectively, $0.406$, $0.293$, and $0.287$
under $\lambda=0.02$, and $0.862$, $0.714$, and $0.654$
under cross-validated $\lambda$. The corresponding ranks
under the fixed ridge are about $33.5$, $64$, and $64$.
Thus the ET gain is partly a positive-rank admissibility gain;
these figures do not identify a pure fixed-rank ET effect.
Signed closed-form weights can also be extremely unstable:
at $m=200$ under cross-validation its mean error ratio is
about $249$, so it is a diagnostic rather than a
structure-preserving competitor.

\begin{figure}[t]
\centering
\includegraphics[width=.91\linewidth]{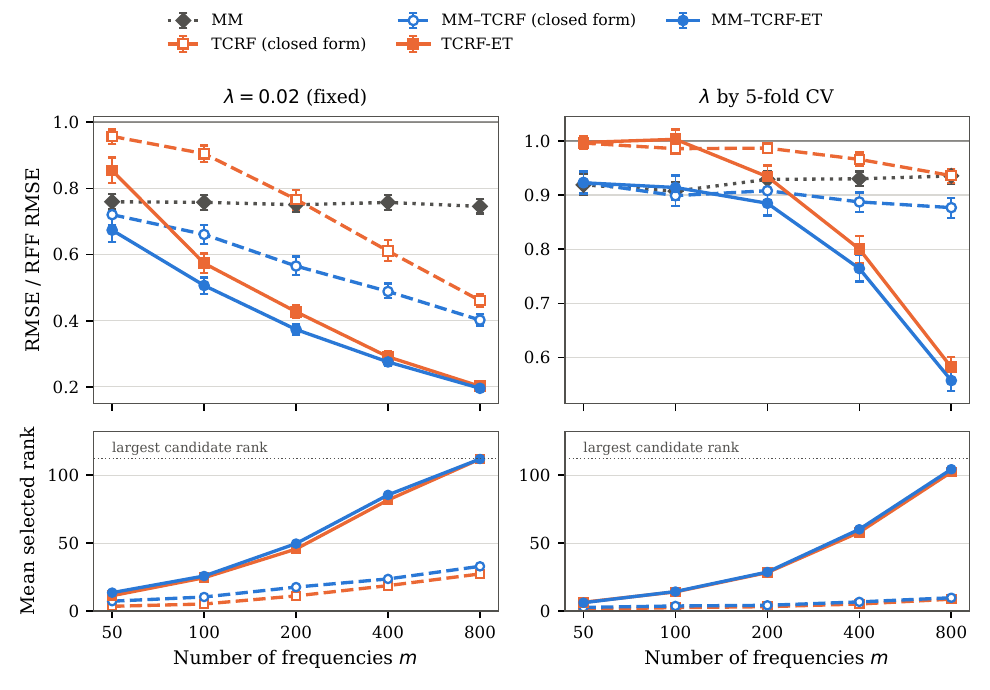}
\caption{Sensitivity of the KRR ET approximation to increasing the
candidate-rank limit from $64$ to $112$. The extended grid uses
an independent $100$-replication run rather than the primary
$200$-replication sample.}
\label{fig:krr-et-extgrid}
\end{figure}

In the independent extended-grid run ($100$ datasets; Figure~\ref{fig:krr-et-extgrid}
and the last column of Table~\ref{tab:krr-rules}), adding
ranks $80$, $96$, and $112$ changes the ET hybrid/RFF ratio
at $m=800$ from $0.287$ to $0.196$ with fixed
$\lambda=0.02$, and from $0.654$ to $0.558$ under
cross-validation. The two replication sets are independent;
these are sensitivity contrasts, not a paired improvement
estimate. The main rank grid often reaches its ceiling $64$,
so the reported rank is not an unconstrained optimum.
Any frequency-equivalent budget estimated by extrapolating
the observed RFF curve past $m=800$ is heuristic
rather than directly observed.

\begin{table}[t]
\centering
\caption{Landmark hybrid under cross-validated full-data ridge (100 paired replications): ratio of mean approximation RMSE to MM. \emph{Full $\lambda$} uses the full-data cross-validated ridge in the landmark subproblem; \emph{landmark CV} selects a separate subproblem ridge by five-fold cross-validation on landmarks. Lower is better.}
\label{tab:landmark-remedy}
\medskip
\small
\begin{tabular}{rlcccccc}
\toprule
 & & \multicolumn{3}{c}{Full $\lambda$} & \multicolumn{3}{c}{Landmark CV} \\
\cmidrule(lr){3-5}\cmidrule(lr){6-8}
$n_L$ & $m$ & CF & ET-pMSE & ET-ESS$_{0.9}$ & CF & ET-pMSE & ET-ESS$_{0.9}$ \\
\midrule
30 & 100 & 0.996 & 1.113 & 0.995 & 1.005 & 1.072 & 1.006 \\
30 & 200 & 0.996 & 1.115 & 1.004 & 0.995 & 1.061 & 1.003 \\
30 & 400 & 0.987 & 1.018 & 0.994 & 0.994 & 1.002 & 0.993 \\
30 & 800 & 0.991 & 0.964 & 0.967 & 0.983 & 0.971 & 0.972 \\
\addlinespace
50 & 100 & 0.999 & 1.094 & 1.003 & 1.000 & 1.059 & 1.000 \\
50 & 200 & 0.994 & 1.122 & 1.000 & 0.990 & 1.052 & 0.995 \\
50 & 400 & 0.975 & 1.008 & 0.978 & 0.982 & 0.979 & 0.979 \\
50 & 800 & 0.975 & 0.954 & 0.953 & 0.974 & 0.935 & 0.935 \\
\addlinespace
100 & 100 & 0.996 & 1.041 & 0.989 & 0.995 & 1.043 & 0.992 \\
100 & 200 & 0.999 & 1.068 & 1.003 & 0.985 & 1.037 & 0.983 \\
100 & 400 & 0.987 & 0.988 & 0.973 & 0.979 & 0.964 & 0.956 \\
100 & 800 & 0.962 & 0.922 & 0.924 & 0.966 & 0.912 & 0.914 \\
\bottomrule
\end{tabular}
\end{table}

The landmark sensitivity check in Table~\ref{tab:landmark-remedy} uses $100$ paired replications, whereas the
primary landmark table (Table~\ref{tab:landmark-et}) has $150$. The
ESS$_{0.9}$ rule admits the largest feasible ET rank with
$m^{-1}\sum_j(w_j^{\mathrm{ET}})^2\le1/0.9$.
This conservative criterion avoids most small-$m$
degradation under the reference-CV ridge; fitting a
separate ridge parameter within the landmark subsample
has relatively limited incremental effect. These results
support an ESS safeguard for the landmark implementation,
not a theorem of monotone predictive improvement.

\subsection{Independent closed-form benchmark and diagnostics}
\label{supp:legacy-closed}

Tables~\ref{tab:krr-detail}--\ref{tab:landmark-detail} report a separate closed-form-only
Monte Carlo benchmark with $\lambda=0.02$ and base seed $20260923$.
These estimates are not paired with the ET/CV tables.
In this KRR and landmark experiment the pilot-standardised
covariance screen and the main-sample correlation check selected
identical ranks; the nonnegative-weight condition was the active
restriction. The largest empirical balance error recorded for
those closed-form fits was below $7\times10^{-16}$.
The KSD recovery diagnostic in Table~\ref{tab:ksd-detail}, which underlies Figure~\ref{fig:ksd-publication}, uses independent subsets and the same base seed.

\begin{table}[t]
\centering
\caption{Detailed kernel ridge regression results. Rank-0 is the percentage of replications in which no positive task-calibration rank was admissible. ESS is the effective sample size of the frequency weights.}
\label{tab:krr-detail}
\medskip
\small
\begin{tabular}{rlrrrrrr}
\toprule
$m$ & Method & Mean RMSE & MCSE & /RFF & Mean rank & Rank-0 (\%) & ESS \\
\midrule
50 & RFF & 0.1083 & 0.0013 & 1.000 & -- & -- & -- \\
50 & MM & 0.0823 & 0.0010 & 0.760 & -- & -- & -- \\
50 & TCRF & 0.1040 & 0.0015 & 0.961 & 3.2 & 41.0 & 47.5 \\
50 & MM--TCRF & 0.0774 & 0.0012 & 0.715 & 7.1 & 13.5 & 46.2 \\
100 & RFF & 0.0774 & 0.0009 & 1.000 & -- & -- & -- \\
100 & MM & 0.0585 & 0.0006 & 0.756 & -- & -- & -- \\
100 & TCRF & 0.0689 & 0.0011 & 0.890 & 6.3 & 19.5 & 93.9 \\
100 & MM--TCRF & 0.0499 & 0.0008 & 0.644 & 11.1 & 2.0 & 92.9 \\
200 & RFF & 0.0546 & 0.0006 & 1.000 & -- & -- & -- \\
200 & MM & 0.0415 & 0.0005 & 0.760 & -- & -- & -- \\
200 & TCRF & 0.0417 & 0.0009 & 0.763 & 11.3 & 5.5 & 189.0 \\
200 & MM--TCRF & 0.0300 & 0.0005 & 0.550 & 17.6 & 0.0 & 187.8 \\
400 & RFF & 0.0389 & 0.0005 & 1.000 & -- & -- & -- \\
400 & MM & 0.0295 & 0.0003 & 0.758 & -- & -- & -- \\
400 & TCRF & 0.0236 & 0.0006 & 0.606 & 18.1 & 3.0 & 380.4 \\
400 & MM--TCRF & 0.0184 & 0.0003 & 0.472 & 24.3 & 0.0 & 381.8 \\
800 & RFF & 0.0273 & 0.0003 & 1.000 & -- & -- & -- \\
800 & MM & 0.0207 & 0.0002 & 0.758 & -- & -- & -- \\
800 & TCRF & 0.0127 & 0.0003 & 0.463 & 26.9 & 0.5 & 770.4 \\
800 & MM--TCRF & 0.0108 & 0.0002 & 0.396 & 31.9 & 0.0 & 774.4 \\
\bottomrule
\end{tabular}
\end{table}

\begin{table}[t]
\centering
\caption{Paired kernel ridge regression gains relative to moment matching. The gain is the RMSE of MM minus that of the method, so positive values favour the method; the positive fraction is the percentage of replications in which the method has smaller approximation RMSE than MM.}
\label{tab:krr-paired}
\medskip
\small
\begin{tabular}{rlrrcr}
\toprule
$m$ & Method & Mean gain & MCSE & 95\% interval & Positive (\%) \\
\midrule
50 & TCRF & $-0.02172$ & 0.00123 & $[-0.02412,-0.01932]$ & 10.5 \\
50 & MM--TCRF & 0.00490 & 0.00078 & $[0.00336,0.00643]$ & 60.0 \\
100 & TCRF & $-0.01041$ & 0.00093 & $[-0.01223,-0.00858]$ & 23.0 \\
100 & MM--TCRF & 0.00863 & 0.00053 & $[0.00760,0.00966]$ & 88.0 \\
200 & TCRF & $-0.00019$ & 0.00075 & $[-0.00166,0.00127]$ & 52.5 \\
200 & MM--TCRF & 0.01148 & 0.00038 & $[0.01073,0.01223]$ & 99.5 \\
400 & TCRF & 0.00590 & 0.00051 & $[0.00490,0.00690]$ & 84.0 \\
400 & MM--TCRF & 0.01111 & 0.00028 & $[0.01055,0.01167]$ & 100.0 \\
800 & TCRF & 0.00806 & 0.00025 & $[0.00756,0.00856]$ & 99.0 \\
800 & MM--TCRF & 0.00989 & 0.00019 & $[0.00951,0.01027]$ & 100.0 \\
\bottomrule
\end{tabular}
\end{table}

\begin{table}[t]
\centering
\caption{Detailed landmark results for the task-calibrated methods. RMSE is the mean RMSE with respect to exact KRR, and Rank-0 is the percentage of replications with no admissible positive rank. The final column is the paired percentage reduction in RMSE relative to MM for the hybrid; the standalone method is left blank in that column.}
\label{tab:landmark-detail}
\medskip
\footnotesize
\begin{tabular}{rrlrrrrrr}
\toprule
$n_L$ & $m$ & Method & RMSE & MCSE & /RFF & Rank & Rank-0 (\%) & Red.\ (\%) \\
\midrule
30 & 100 & Landmark-TCRF & 0.0708 & 0.0010 & 0.940 & 6.7 & 6.7 & -- \\
30 & 100 & MM--Landmark-TCRF & 0.0565 & 0.0007 & 0.750 & 11.1 & 0.7 & 1.6 \\
30 & 200 & Landmark-TCRF & 0.0474 & 0.0008 & 0.885 & 10.8 & 2.0 & -- \\
30 & 200 & MM--Landmark-TCRF & 0.0383 & 0.0005 & 0.716 & 15.1 & 0.0 & 5.8 \\
30 & 400 & Landmark-TCRF & 0.0308 & 0.0005 & 0.817 & 16.5 & 0.7 & -- \\
30 & 400 & MM--Landmark-TCRF & 0.0252 & 0.0004 & 0.668 & 21.1 & 0.0 & 10.7 \\
30 & 800 & Landmark-TCRF & 0.0196 & 0.0003 & 0.742 & 22.6 & 0.0 & -- \\
30 & 800 & MM--Landmark-TCRF & 0.0169 & 0.0002 & 0.637 & 25.3 & 0.0 & 16.3 \\
50 & 100 & Landmark-TCRF & 0.0689 & 0.0011 & 0.914 & 7.1 & 7.3 & -- \\
50 & 100 & MM--Landmark-TCRF & 0.0546 & 0.0008 & 0.726 & 11.1 & 1.3 & 4.9 \\
50 & 200 & Landmark-TCRF & 0.0459 & 0.0008 & 0.857 & 10.7 & 1.3 & -- \\
50 & 200 & MM--Landmark-TCRF & 0.0364 & 0.0005 & 0.681 & 15.5 & 0.0 & 10.6 \\
50 & 400 & Landmark-TCRF & 0.0286 & 0.0005 & 0.757 & 17.1 & 0.0 & -- \\
50 & 400 & MM--Landmark-TCRF & 0.0234 & 0.0003 & 0.620 & 21.5 & 0.0 & 17.1 \\
50 & 800 & Landmark-TCRF & 0.0183 & 0.0003 & 0.692 & 22.8 & 0.0 & -- \\
50 & 800 & MM--Landmark-TCRF & 0.0156 & 0.0002 & 0.588 & 25.8 & 0.0 & 22.8 \\
100 & 100 & Landmark-TCRF & 0.0679 & 0.0011 & 0.902 & 6.9 & 6.0 & -- \\
100 & 100 & MM--Landmark-TCRF & 0.0524 & 0.0008 & 0.696 & 11.1 & 2.0 & 8.9 \\
100 & 200 & Landmark-TCRF & 0.0439 & 0.0008 & 0.821 & 11.1 & 0.7 & -- \\
100 & 200 & MM--Landmark-TCRF & 0.0340 & 0.0005 & 0.636 & 16.1 & 0.0 & 16.5 \\
100 & 400 & Landmark-TCRF & 0.0260 & 0.0005 & 0.690 & 16.9 & 0.0 & -- \\
100 & 400 & MM--Landmark-TCRF & 0.0210 & 0.0003 & 0.558 & 22.5 & 0.0 & 25.7 \\
100 & 800 & Landmark-TCRF & 0.0156 & 0.0003 & 0.590 & 23.7 & 0.0 & -- \\
100 & 800 & MM--Landmark-TCRF & 0.0137 & 0.0002 & 0.517 & 26.0 & 0.0 & 31.9 \\
\bottomrule
\end{tabular}
\end{table}

\begin{table}[t]
\centering
\caption{KSD subset-recovery diagnostics with independent pilot subsets. The two sweeps vary $L$ at fixed $s=50$ and vary $s$ at fixed $L=30$, respectively.}
\label{tab:ksd-detail}
\medskip
\small
\begin{tabular}{llcrrr}
\toprule
Scenario & Sweep & Setting & Mean $\rho_D$ & MCSE & 95\% interval \\
\midrule
Mixture & $L$ & $L=1,\ s=50$ & 0.9287 & 0.0066 & [0.9157,0.9416] \\
Mixture & $L$ & $L=3,\ s=50$ & 0.9023 & 0.0072 & [0.8881,0.9165] \\
Mixture & $L$ & $L=10,\ s=50$ & 0.8484 & 0.0077 & [0.8334,0.8634] \\
Mixture & $L$ & $L=30,\ s=50$ & 0.7214 & 0.0095 & [0.7029,0.7400] \\
Mixture & $L$ & $L=100,\ s=50$ & 0.4667 & 0.0082 & [0.4505,0.4829] \\
Mixture & $s$ & $L=30,\ s=25$ & 0.8687 & 0.0076 & [0.8538,0.8837] \\
Mixture & $s$ & $L=30,\ s=50$ & 0.7186 & 0.0094 & [0.7002,0.7371] \\
Mixture & $s$ & $L=30,\ s=100$ & 0.4077 & 0.0079 & [0.3923,0.4231] \\
Mixture & $s$ & $L=30,\ s=200$ & 0.1411 & 0.0036 & [0.1340,0.1482] \\
$t_5$ & $L$ & $L=1,\ s=50$ & 0.4884 & 0.0053 & [0.4780,0.4988] \\
$t_5$ & $L$ & $L=3,\ s=50$ & 0.4510 & 0.0054 & [0.4404,0.4615] \\
$t_5$ & $L$ & $L=10,\ s=50$ & 0.3546 & 0.0052 & [0.3444,0.3649] \\
$t_5$ & $L$ & $L=30,\ s=50$ & 0.2181 & 0.0040 & [0.2103,0.2259] \\
$t_5$ & $L$ & $L=100,\ s=50$ & 0.0961 & 0.0021 & [0.0920,0.1002] \\
$t_5$ & $s$ & $L=30,\ s=25$ & 0.3672 & 0.0053 & [0.3569,0.3776] \\
$t_5$ & $s$ & $L=30,\ s=50$ & 0.2231 & 0.0040 & [0.2153,0.2309] \\
$t_5$ & $s$ & $L=30,\ s=100$ & 0.0966 & 0.0020 & [0.0926,0.1005] \\
$t_5$ & $s$ & $L=30,\ s=200$ & 0.0325 & 0.0007 & [0.0312,0.0337] \\
\bottomrule
\end{tabular}
\end{table}

\subsection{Exact-KSD benchmark for both subset sizes and timing}
\label{supp:ksd-exact}

\begin{figure}[t]
\centering
\includegraphics[width=.94\linewidth]{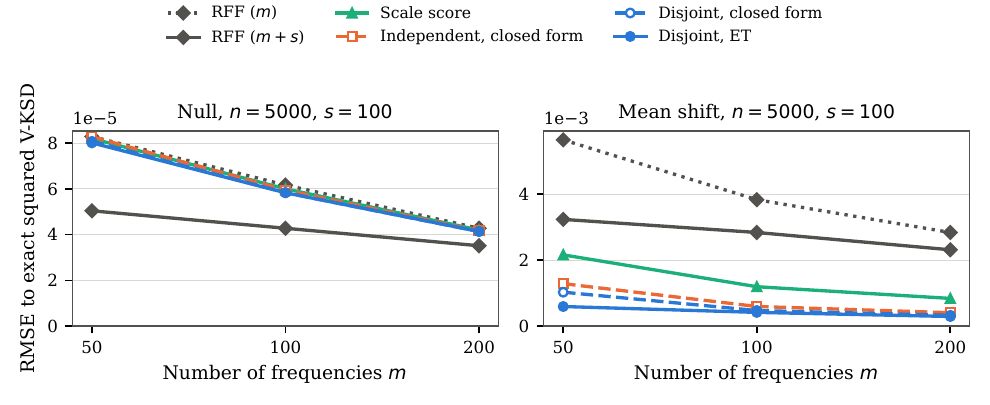}
\caption{Absolute RMSE to the exact empirical squared V-KSD as a function of the
number of frequencies $m$ at $n=5000$ and $s=100$. Left: Gaussian null; right:
fixed mean-shift alternative.}
\label{fig:ksd-exact-frequency}
\end{figure}

Table~\ref{tab:ksd-exact-grid} extends Table~\ref{tab:ksd-et} to both
subset sizes $s\in\{50,100\}$ and all frequency budgets $m\in\{50,100,200\}$,
using the same run: target $p=N(0,I_{10})$, Gaussian bandwidth
$\sigma=\sqrt{10}$, data from $N(0,I_{10})$ or $N(0.5e_1,I_{10})$,
$n\in\{1000,2000,5000\}$, 100 datasets for each scenario and $n$, eight
frequency draws per dataset, and seed $20261024$. The exact squared V-KSD
uses the analytic Stein kernel on every ordered pair, evaluated by block
matrix products to avoid storing an $n\times n$ array. For each dataset
$D_b$, its exact squared V-KSD $\theta_{D_b}$ is held fixed across frequency
replicates $j=1,\ldots,8$, and the reported accuracy statistic is
\[
\operatorname{RMSE}(\mathrm{method})
=\left(\frac{1}{100\cdot8}\sum_{b=1}^{100}\sum_{j=1}^{8}
\bigl(\hat\theta_{b,j}^{\mathrm{method}}-\theta_{D_b}\bigr)^2\right)^{1/2}.
\]
Pilot subsets, with $L=n/s$, are drawn once per dataset, and every value of
$m$ uses a prefix of one frequency sample of size $300$, so that comparisons
across methods and frequency budgets are paired. Closed-form weights use
candidate ranks $2,1,0$, in this order; rank one contains only the centred
scale score $q(\omega)=\|\omega\|^2-10/\sigma^2$, and rank two adds the
averaged, exactly centred subset frequency integrand. At the selected rank
the minimum weight must be at least $0.01$ and the standardised
sample-covariance eigenvalue ratio at least $10^{-9}$; at rank zero the
estimator reduces to ordinary RFF. ET and closed-form pilots use the same
controls, and KSD ET chooses the largest converged admissible rank rather
than the KRR pilot-proxy rule. Exact subset centring costs $Ls^2=ns$
kernel-pair evaluations in both pilot designs, comparable to $s$ additional
frequencies, which defines the RFF ($m+s$) column.
Figure~\ref{fig:ksd-exact-frequency} shows the corresponding absolute errors at
$n=5000$ and $s=100$. Under the null both positive calibration solutions have
nearly the same RMSE and are less accurate than RFF ($m+s$), while under the
mean shift ET improves substantially at $m=50$ over the positivity-screened
closed-form calibration and both remain far more accurate than RFF ($m+s$).
No significance-test size or power is inferred from these computational
approximation experiments.

\begin{table}[t]
\centering
\caption{Exact empirical squared V-KSD benchmark for both subset sizes: ratio of frequency-approximation RMSE to ordinary RFF with $m$ frequencies, over 100 datasets and eight frequency draws per dataset. RFF ($m+s$) spends the pilot budget of $ns$ kernel evaluations on $s$ additional frequencies; CF and ET denote closed-form and exponential-tilting weights with independent (Indep.) or disjoint-block (Disj.) pilots. The rows with $s=100$, $m\in\{50,200\}$ repeat Table~\ref{tab:ksd-et}.}
\label{tab:ksd-exact-grid}
\medskip
\footnotesize
\setlength{\tabcolsep}{4pt}
\begin{tabular}{llrrccccc}
\toprule
Scenario & $n$ & $s$ & $m$ & RFF ($m+s$) & Indep.\ CF & Indep.\ ET & Disj.\ CF & Disj.\ ET \\
\midrule
Null & 1000 & 50 & 50 & 0.698 & 0.972 & 0.974 & 0.992 & 0.999 \\
Null & 1000 & 50 & 100 & 0.822 & 0.952 & 0.951 & 0.971 & 0.966 \\
Null & 1000 & 50 & 200 & 0.891 & 0.961 & 0.960 & 0.959 & 0.959 \\
Null & 1000 & 100 & 50 & 0.574 & 0.959 & 0.961 & 0.962 & 0.957 \\
Null & 1000 & 100 & 100 & 0.717 & 0.941 & 0.936 & 0.956 & 0.955 \\
Null & 1000 & 100 & 200 & 0.810 & 0.924 & 0.923 & 0.938 & 0.938 \\
Null & 2000 & 50 & 50 & 0.720 & 0.975 & 0.973 & 0.982 & 0.985 \\
Null & 2000 & 50 & 100 & 0.802 & 0.968 & 0.967 & 0.964 & 0.963 \\
Null & 2000 & 50 & 200 & 0.875 & 0.960 & 0.960 & 0.956 & 0.955 \\
Null & 2000 & 100 & 50 & 0.578 & 0.961 & 0.949 & 0.978 & 0.971 \\
Null & 2000 & 100 & 100 & 0.713 & 0.956 & 0.955 & 0.953 & 0.951 \\
Null & 2000 & 100 & 200 & 0.806 & 0.951 & 0.950 & 0.941 & 0.939 \\
Null & 5000 & 50 & 50 & 0.744 & 0.990 & 0.988 & 0.974 & 0.975 \\
Null & 5000 & 50 & 100 & 0.817 & 0.963 & 0.962 & 0.969 & 0.969 \\
Null & 5000 & 50 & 200 & 0.905 & 0.971 & 0.971 & 0.974 & 0.975 \\
Null & 5000 & 100 & 50 & 0.608 & 0.999 & 0.994 & 0.974 & 0.967 \\
Null & 5000 & 100 & 100 & 0.693 & 0.964 & 0.958 & 0.946 & 0.945 \\
Null & 5000 & 100 & 200 & 0.821 & 0.974 & 0.974 & 0.967 & 0.967 \\
\midrule
Mean shift & 1000 & 50 & 50 & 0.734 & 0.353 & 0.272 & 0.339 & 0.253 \\
Mean shift & 1000 & 50 & 100 & 0.812 & 0.285 & 0.277 & 0.255 & 0.245 \\
Mean shift & 1000 & 50 & 200 & 0.915 & 0.277 & 0.275 & 0.242 & 0.237 \\
Mean shift & 1000 & 100 & 50 & 0.596 & 0.334 & 0.251 & 0.293 & 0.210 \\
Mean shift & 1000 & 100 & 100 & 0.712 & 0.248 & 0.243 & 0.223 & 0.196 \\
Mean shift & 1000 & 100 & 200 & 0.824 & 0.248 & 0.245 & 0.195 & 0.193 \\
Mean shift & 2000 & 50 & 50 & 0.687 & 0.318 & 0.228 & 0.311 & 0.207 \\
Mean shift & 2000 & 50 & 100 & 0.835 & 0.274 & 0.234 & 0.224 & 0.214 \\
Mean shift & 2000 & 50 & 200 & 0.904 & 0.237 & 0.232 & 0.216 & 0.210 \\
Mean shift & 2000 & 100 & 50 & 0.574 & 0.299 & 0.218 & 0.267 & 0.167 \\
Mean shift & 2000 & 100 & 100 & 0.710 & 0.218 & 0.208 & 0.182 & 0.164 \\
Mean shift & 2000 & 100 & 200 & 0.824 & 0.211 & 0.207 & 0.169 & 0.162 \\
Mean shift & 5000 & 50 & 50 & 0.679 & 0.262 & 0.167 & 0.231 & 0.150 \\
Mean shift & 5000 & 50 & 100 & 0.844 & 0.175 & 0.165 & 0.161 & 0.152 \\
Mean shift & 5000 & 50 & 200 & 0.875 & 0.166 & 0.157 & 0.150 & 0.141 \\
Mean shift & 5000 & 100 & 50 & 0.573 & 0.228 & 0.142 & 0.182 & 0.105 \\
Mean shift & 5000 & 100 & 100 & 0.741 & 0.155 & 0.143 & 0.123 & 0.108 \\
Mean shift & 5000 & 100 & 200 & 0.815 & 0.143 & 0.138 & 0.115 & 0.103 \\
\bottomrule
\end{tabular}
\end{table}

A separate timing check measured wall-clock times on a two-thread CPU at
$n=5000$, $s=100$ and $m=200$ under the mean shift, taking the minimum of
three runs for each of five datasets. The mean times were $0.652$ seconds for
the exact squared V-KSD, $0.088$ seconds for the disjoint-block calibrated
approximation with closed-form weights and $0.087$ seconds with ET weights,
both including the exact pilot centring and the main features, and $0.038$
and $0.060$ seconds for ordinary RFF with $m$ and $m+s$ frequencies.
Pilot-subset index generation and frequency generation are excluded from the
timed paths. In wall-clock time the pilot therefore costs about as much as
$230$ additional frequencies, and the calibrated approximation is about seven
times faster than the exact statistic.

\section{Pilot-inclusive large-sample KRR checks}
\label{sec:supp-large-krr}

This section isolates the large-sample behaviour of the
\emph{closed-form} NY--MM--TCRF construction, which combines
Nystr\"om-approximated influence coefficients, moment matching and TCRF
calibration, under its fixed $\lambda=0.001$ protocol. It is an additional numerical check rather
than a comparison between ET and closed-form weighting or between
ridge-tuning procedures. The data-generating model matches the
Gaussian KRR model of Section~\ref{sec:numerical}.

The large-sample experiment uses $n=10{,}000$, $d=5$, independent test
samples of size $2000$, Gaussian-kernel bandwidth obtained from a
$600$-observation training subsample, and ridge parameter $\lambda=0.001$
under the $K+n\lambda I$ convention. To avoid transductive design, the $J$
prediction-influence design points (the representative points of
Section~\ref{sec:krr-main}) are generated independently from
$N(0,I_5)$, without using test responses or test covariates.
The pilot uses $30$ training landmarks in a full-sample Nystr\"om KRR
approximation, $M_0=200$ pilot frequencies and principal task directions;
there are no additional inducing points for centring. A degree-four
Gaussian Taylor kernel approximation provides $126$ explicit centring
features. We use ranks $\{2,4,6,8,12,16,20,24\}$, minimum admissible
weight $0.01$ and main-sample correlation conditioning threshold $10^{-4}$.
A separate one-worker timing check used $12$ replications per frequency
budget. Mean fitting times (NY--MM--TCRF versus ORF) were $1.269$ versus
$1.206$ seconds at $m=750$ and $1.368$ versus $1.459$ seconds at
$m=800$. Bootstrap 95\% intervals for the corresponding ratios of mean
times were $[0.998,1.107]$ and $[0.864,1.020]$. These times include
pilot construction, Taylor centring, feature matrix construction,
calibration and the ridge solve, but exclude data generation, the
reference fit, bandwidth selection and frequency generation.
The intervals include unity, so the measured runtimes do not establish
an equal-time advantage.

The benchmark is a separately fitted 500-inducing-point Nystr\"om KRR,
not full exact KRR. Orthogonal random features (ORF) use frequencies generated in
independent blocks of $d$ orthogonal directions with $\chi_d$ radial corrections.
ORF has $m+M_0$ frequencies, whereas NY--MM--TCRF has $m$ prediction
frequencies and $M_0$ frequencies used only for task pilot construction.

\begin{table}[t]
\centering
\caption{Large-sample closed-form KRR approximation with the
pilot-inclusive frequency budget ($n=10{,}000$, $M_0=200$, $100$
paired replications). RMSE is measured against the 500-point
Nystr\"om KRR reference. The last column reports the ratio of
mean RMSEs with a 95\% paired-bootstrap interval.}
\label{tab:supp-large-main}
\medskip
\footnotesize
\setlength{\tabcolsep}{3pt}
\begin{tabular}{ccccc}
\toprule
$m$ & ORF ($m+200$) & NY--MM--TCRF ($J=20$) &
NY--MM--TCRF ($J=80$) & $J=80$/ORF \\
\midrule
700 & 0.03196 & 0.03229 & 0.03153 & 0.987 [0.956,1.019] \\
750 & 0.03184 & 0.03007 & \textbf{0.02966} &
\textbf{0.931 [0.903,0.961]} \\
800 & 0.03129 & 0.02948 & \textbf{0.02870} &
\textbf{0.917 [0.887,0.950]} \\
\bottomrule
\end{tabular}
\end{table}

At $m=750$ and $800$, the $J=80$ closed-form method reduces
frequency-approximation RMSE relative to the pilot-budget-matched
ORF reference by $6.9\%$ and $8.3\%$, respectively; at $m=700$
the difference is not resolved. Table~\ref{tab:supp-large-grid} reports the
complete pilot-inclusive frequency grid for $J=20$. This does not establish
superiority in predictive risk against the true regression
function or on an equal-wall-time basis.

\begin{table}[t]
\centering
\caption{Complete pilot-inclusive frequency grid with $J=20$ design points.
Each entry is the mean RMSE over $100$ paired replications relative to the
independently fitted Nystr\"om-500 reference. The last column is a
ratio of mean RMSEs with a 95\% paired-bootstrap interval.}
\label{tab:supp-large-grid}
\medskip
\small
\begin{tabular}{cccc}
\toprule
$m$ & ORF ($m+200$) & NY--MM--TCRF ($J=20$) & TCRF/ORF [95\% CI] \\
\midrule
400 & 0.03959 & 0.04272 & 1.079 [1.046,1.113] \\
500 & 0.03700 & 0.03685 & 0.996 [0.964,1.031] \\
600 & 0.03453 & 0.03428 & 0.993 [0.962,1.025] \\
700 & 0.03196 & 0.03229 & 1.011 [0.980,1.042] \\
750 & 0.03184 & 0.03007 & 0.945 [0.916,0.974] \\
800 & 0.03129 & 0.02948 & 0.942 [0.913,0.973] \\
\bottomrule
\end{tabular}
\end{table}

For $m=700$, we explored $J=10,20,40,80$ and four independent seeds for
$J=20$ on the first $50$ replications while holding the training data,
main/pilot frequencies and pilot landmarks fixed. The $J=80$ choice was
then validated on the remaining $50$ replications. The resulting ratios
of mean RMSE for $J=80$ relative to $J=20$ were $0.978$ on the exploratory
half and $0.975$ on the held-out half (95\% paired-bootstrap interval
$[0.959,0.990]$ on the latter). With $100$ replications in total, the
mean selected ranks for $J=80$ and $J=20$ were $15.94$ and $13.70$, respectively. Changing the
design point count alters the available principal task space, so this is
a model-specific tuning effect rather than a monotonicity theorem.
The $m=750,800$ results for $J=80$ in Table~\ref{tab:supp-large-main}
were computed using
different data seeds from those used to select $J$ at $m=700$.

For the $J=80$ runs with fresh data seeds, the mean NY--MM--TCRF RMSE
(Monte Carlo SE) was $0.02966$ ($0.00033$) at $m=750$ and $0.02870$
($0.00031$) at $m=800$, with mean selected ranks $17.02$ and $16.86$.
The corresponding 100-replication ORF means were $0.03184$ and $0.03129$;
the ratios of means and their paired-bootstrap intervals are reported
in Table~\ref{tab:supp-large-main}. On the same paired samples the RMSE ratios
of $J=80$ relative to $J=20$ were $0.986$ ($[0.971,1.001]$) at
$m=750$ and $0.973$ ($[0.960,0.986]$) at $m=800$.

The small-landmark influence defined in Section~\ref{sec:krr-main} is exactly centred because its kernel
expectations are computed on the landmark set. The large-sample
implementation instead approximates coefficients of the full-data
influence and may also approximate the kernel used to centre it.
Let $\tilde\alpha$ and $\tilde t_x$ approximate
$\alpha=(K+n\lambda I_n)^{-1}y$ and
$(K+n\lambda I_n)^{-1}k_x$, respectively, and let $K_p$ be the
degree-$p$ Taylor approximation of $K$. The implemented influence is
\[
\tilde b_{x,p}(Z)
=(a_Z(x)-k_x)^{\top}\tilde\alpha
-\tilde t_x^{\top}(A_Z-K_p)\tilde\alpha .
\]
Because $\mathbb{E}_{q_0}[a_Z(x)]=k_x$ and
$\mathbb{E}_{q_0}[A_Z]=K$, its spectral mean is
$\mathbb{E}_{q_0}[\tilde b_{x,p}(Z)]=\tilde t_x^{\top}(K_p-K)\tilde\alpha$.
Thus approximate Nystr\"om coefficients alone preserve exact centring
when the kernel expectation $K$ is retained, whereas Taylor centring
can introduce a nonzero spectral mean.
For the implemented calibration vector $\tilde c_p$, write
$\delta_p=\mathbb{E}_{q_0}[\tilde c_p(Z)]$. If the weights are
normalised to $m^{-1}\sum_jw_j=1$, exact empirical balance gives
\[
\frac1m\sum_{j=1}^m w_j\tilde c_p(Z_j)=0,
\quad
\frac1m\sum_{j=1}^m w_j
(\tilde c_p(Z_j)-\delta_p)=-\delta_p .
\]
Hence exact empirical balance and exact population centring are distinct;
a fixed Taylor degree does not by itself make $\delta_p$ vanish as
$m\to\infty$, whereas the fixed-rank projection and second-order theorems
of Section~\ref{sec:main} assume exactly centred calibration functions. To quantify
the centring error, condition on the training dataset, the calibration
directions and the selected rank $r$, let $\tilde c_{p,k}$ be the $k$th implemented control,
and put $\delta_{p,k}=\mathbb{E}_{q_0}[\tilde c_{p,k}(Z)]$ and
\[
D_m=\max_{1\le k\le r}
\frac{\sqrt m\,|\delta_{p,k}|}{\hat s_k},
\]
where $\hat s_k$ is the empirical standard deviation of the $k$th
implemented control over the prediction frequencies, with $D_m=0$ when
$r=0$. An independent diagnostic experiment with 100 datasets and three
frequency budgets evaluated $D_m$ for a Taylor-centred variant that uses a
Nystr\"om influence with $100$ landmarks, principal directions estimated from
the prediction frequencies and exponential-tilting weights. Over its $300$
fits the median and maximum of $D_m$ were $0.30$ and $0.69$. These diagnostics belong to that
separate experiment and are not matched to the RMSE entries of
Table~\ref{tab:supp-large-main}; $D_m$ is a standardised diagnostic of
centring error, not a confidence bound, and the moment-matched frequencies
and the data-adaptive selection of directions preclude interpreting
$\hat s_k/\sqrt m$ as an automatically valid sampling standard error.

\section{Derivation of the estimated-sampler expansion}
\label{supp:estimated-sampler}

We derive the first-order expansion used in Section~\ref{sec:estimated-sampler}. Retain its notation: $X_1,\ldots,X_m$ are independent from
$p(x;\eta_0)$, $u(X)=f(X)/p(X;\eta_0)-\theta$, and
$s(X)=\partial_\eta\log p(X;\eta_0)$. Assume a common support,
an interior true parameter, nonsingular Fisher information
$G=\mathbb{E}[s(X)s(X)^\top]$, and the usual differentiability,
integrability, and likelihood regularity conditions.

Since $\hat\eta$ is the MLE computed from these same observations,
Taylor expansion of its score equation and the information identity give
\[
0=\frac1m\sum_{j=1}^m s(X_j;\hat\eta)
=\frac1m\sum_{j=1}^m s(X_j)
-(G+o_p(1))(\hat\eta-\eta_0),
\]
and hence
\begin{equation}
\sqrt m(\hat\eta-\eta_0)
=G^{-1}\frac1{\sqrt m}\sum_{j=1}^m s(X_j)+o_p(1).
\label{eq:supp-es-mle}
\end{equation}

Differentiating the importance weight with the observations held fixed,
\[
\left.\frac{\partial}{\partial\eta^\top}
\frac{f(x)}{p(x;\eta)}\right|_{\eta=\eta_0}
=-\frac{f(x)}{p(x;\eta_0)}s(x)^\top.
\]
Because $f(X)/p(X;\eta_0)=u(X)+\theta$ and
$\mathbb{E}[s(X)]=0$, the expectation of this derivative is
$-\mathbb{E}[u(X)s(X)^\top]$. A Taylor expansion of the weighted
average around $\eta_0$ therefore yields
\begin{equation}
\sqrt m(\hat\theta-\theta)
=\frac1{\sqrt m}\sum_{j=1}^m u(X_j)
-\mathbb{E}[u(X)s(X)^\top]\sqrt m(\hat\eta-\eta_0)
+o_p(1).
\label{eq:supp-es-weight}
\end{equation}
The remainder in \eqref{eq:supp-es-weight} follows from the assumed
smoothness and a uniform law of large numbers for the derivative near
$\eta_0$. Substituting \eqref{eq:supp-es-mle} gives
\begin{equation}
\sqrt m(\hat\theta-\theta)
=\frac1{\sqrt m}\sum_{j=1}^m
\left(u(X_j)-\mathbb{E}[u(X)s(X)^\top]G^{-1}s(X_j)\right)
+o_p(1).
\label{eq:supp-es-final}
\end{equation}
The subtracted term is the $L_2(p_{\eta_0})$ projection of $u$
onto the score span. Thus the $\sqrt m$-scale asymptotic variance is
\[
\operatorname{avar}(\hat\theta)
=\mathbb{E}[u(X)^2]
-\mathbb{E}[u(X)s(X)^\top]G^{-1}\mathbb{E}[s(X)u(X)].
\]
The covariance cancellation in \eqref{eq:supp-es-final} relies on using
the same sample for $\hat\eta$ and $\hat\theta$. Although
$\mathbb{E}_{\eta}[f(X)/p(X;\eta)]=\theta$ is constant in $\eta$,
the derivative above keeps the sampling distribution fixed at $\eta_0$.

\end{document}